\documentclass[11pt]{article}

\usepackage[preprint]{acl}
\usepackage{times}
\usepackage{latexsym}
\usepackage[T1]{fontenc}
\usepackage[utf8]{inputenc}
\usepackage{microtype}
\usepackage{graphicx}
\usepackage{amsmath,amssymb}
\usepackage{booktabs}
\usepackage{multirow}
\usepackage{array}
\usepackage{tablefootnote}
\usepackage{threeparttable}
\usepackage{placeins}
\usepackage{algpseudocode}

\newcommand{\hide}[1]{}
\newcommand{\tas}{\textsc{trace as state}}

\newcommand{\tappend}{\textsc{trace append}}
\newcommand{\ntr}{n_{\mathrm{tr}}}

\newcommand{\inputhook}[1]{%
  \IfFileExists{#1.tex}{%
    \input{#1}%
  }{%
    \typeout{Missing input hook: #1.tex}%
  }%
}

\title{Trace as State: Reasoning Traces as Conditional States for Long-Context Transformers}
\author{Xu Zou$^{1}$ \quad Jie Tang$^{2}$ \\
$^1$Z.ai\quad $^2$Tsinghua University
}
\date{\today}

\begin{document}

\maketitle

\begin{abstract}
Transformers process information causally, but long-context
reasoning may depend on task state discovered only later. We formalize this
mismatch through conditional state update tasks. For causal state update processors, providing the condition first can require
exponentially less memory in the worst case than providing it last.

Motivated by this principle, we introduce \tas{}. We use collected reasoning traces as a textual proxy for task
state and place it before the long-context block on a fresh pass, allowing
information derived previously to guide rereading. 

We conduct extensive experiments on \tas{} and \tappend{}, a matched control that uses the same task state proxy but put it after the context. 
Across three models and three long-context datasets, \tas{} outperforms
\tappend{} in 26 of 27 reported combinations of model, task, and metric. On
GraphWalks Parents, exact match lifts DeepSeek V4 Pro Preview from 29.2\% on the initial pass and 43.0\%
with \tappend{} to 81.8\% with \tas{}, and from 66.4\% and
83.2\% to 100.0\% for GLM-5.2. These results show that placing traces before the context can improve long-context reasoning while retaining the
causal transformer structure.
\end{abstract}

\section{Introduction}

Frontier language models can now accept inputs extending to hundreds of
thousands or even millions of tokens. Advances in sparse, compressed, and
hybrid sequence modeling architectures have substantially enlarged their
nominal context windows
\cite{deepseekai2025deepseekv32,deepseekai2026deepseekv4,
qwen2025qwen3next,yang2025gateddelta}. 

Despite these architectural advances, autoregressive Transformers retain
causal attention. Information appearing later remains available to subsequent
reasoning and answer generation, but it cannot affect the representations
formed at earlier positions within the same pass.

One structural difficulty is a mismatch between input order and reasoning
order. Prior work shows that reasoning performance can depend not only on
where supporting information appears, but also on the relative order in which
it is presented \cite{chen2024premiseorder,yu2025core}. In many long context
tasks, a solver must maintain task-relevant state, such as an active target, a
search frontier, or a set of rejected hypotheses. Some of this state may
become available only after earlier parts of the context have already been
processed. 

\begin{figure*}[!t]
  \centering
  \includegraphics[width=\textwidth]{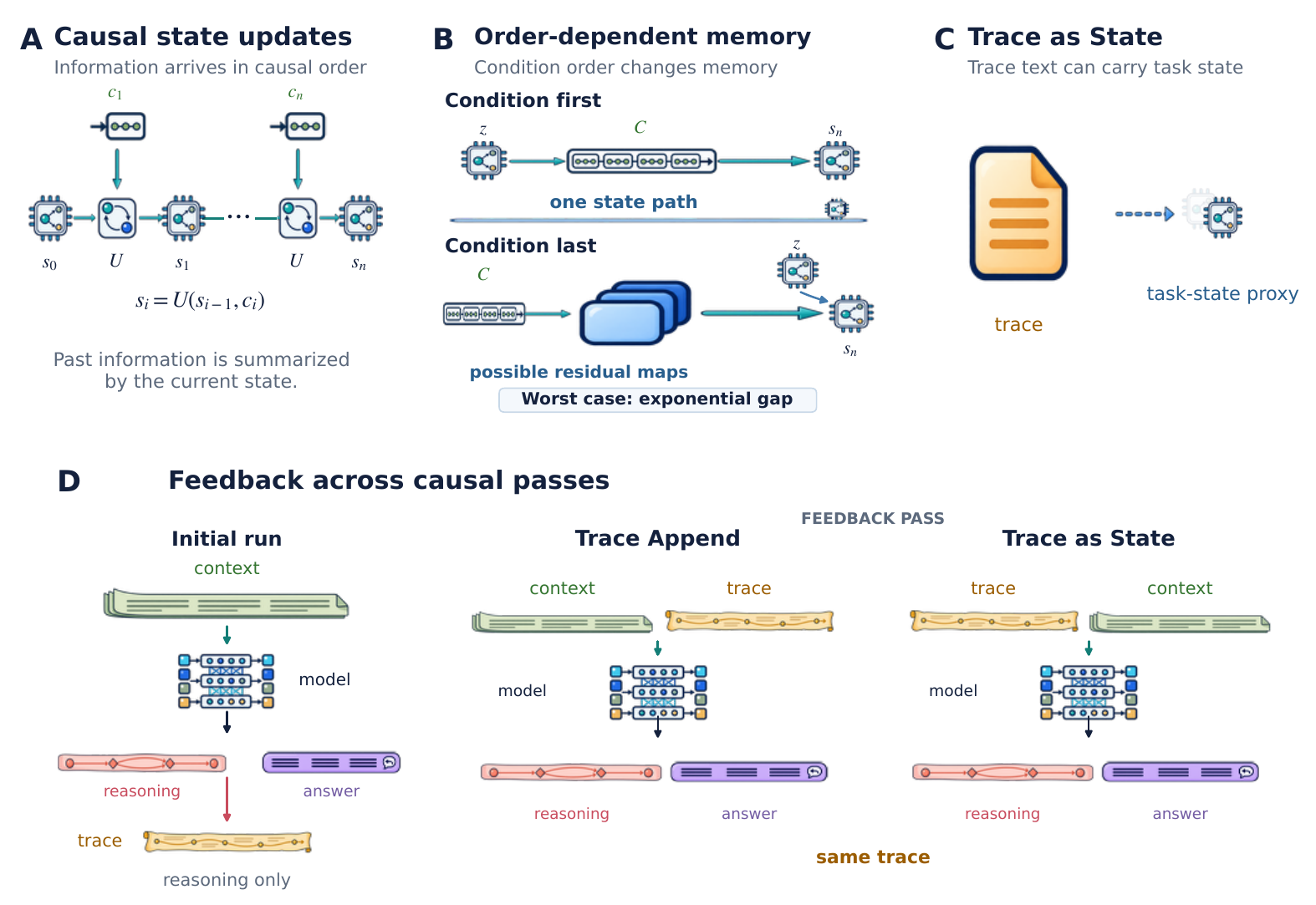}
  \caption{\textbf{Overview for \tas{}}
  (A) A causal processor carries one running state.
  (B) In a worst-case conditional state update task, condition-first
  processing tracks only the realized state, whereas condition-last processing
  can require exponentially more working memory.
  (C) Reasoning trace $T$ provides a textual proxy for task state.
  (D) \tas{} places $T$ before the context in a fresh causal pass. }
  \label{fig:trace-as-state-overview}
\end{figure*}

We analyze this asymmetry through conditional state update tasks. Such a task
begins from a state specified by a condition and applies an information
sequence to it. If the condition is
available before the sequence, a causal processor can update the realized
state as each item arrives. If the condition arrives after the sequence,
the processor may instead need to retain how the sequence would act on every
possible condition. We show that these two input orders can have exponentially
different memory requirements in the worst case. The separation
motivates an ordering principle for causal transformer passes: task-relevant
state discovered late in one pass can be made available before the context in
the next pass.
Figure~\ref{fig:trace-as-state-overview}A and ~\ref{fig:trace-as-state-overview}B illustrate the
causal processor and the two input orders.

Modern reasoning models provide an
observable interface through the reasoning traces they generate before
producing visible answers
\cite{deepr1,qwen2025qwen3technical,criticalthinking}.
Generated reasoning text can carry intermediate computational information
across steps
\cite{scratchpad,cot,merrill2024expressive,stateovertokens}.
Reasoning traces may be incomplete or contain errors, and we treat them as an
observable textual proxy for task state, as shown in
figure~\ref{fig:trace-as-state-overview}C. We therefore introduce \tas{}, a
general inference scaling method that follows a read, compute, feedback, and
reread procedure. The model first processes the task and generates one or more
reasoning traces. We serialize the collected traces as $T$ and place it before
the long-context block in a fresh pass. 

We compare \tas{} with \tappend{}, a placement control that gives the second
pass serialized trace text after the long context.
Under \tas{}, the trace is available while the context is processed again.
Under \tappend{}, the trace can still guide later reasoning and answer
generation, but it cannot influence the representations already formed for
the preceding context. Figure~\ref{fig:trace-as-state-overview}D
summarizes the strict placement design.

We evaluate this placement intervention on GraphWalks 256K, MRCRv2 8-needle,
and NUB-1M using three frontier models from different providers: DeepSeek V4
Pro Preview, GLM-5.2, and Qwen 3.7 Max.
Across the 27 reported combinations of model, task, and metric, \tas{}
outperforms \tappend{} in 26. The result shows that the benefit of using reasoning traces as a task state proxy and placing it before the context is general in long context reasoning, and is consistent with our conditional state update task memory analysis. 

We further test alternative explanations through ablations on GraphWalks 256K
using DeepSeek V4 Pro Preview. These controls weaken explanations based solely on
generic trace scaffolding or access to first-pass answers and further support
the \tas{} interpretation.

Our contributions are as follows:
\vspace{-0.05in}
\begin{itemize}
  \item We use a conditional state update abstraction to characterize a
  worst-case exponential order separation for deterministic processors that
  read the input once,
  and apply it as a qualitative principle for causal
  processing of long contexts.
  \item We introduce \tas{}, a general inference scaling method for reasoning
  over long contexts that feeds reasoning traces back as textual state
  proxies. The method adds feedback between passes while retaining causal
  processing within each pass.
  \item Across three models and three benchmarks for reasoning over long
  contexts, \tas{} scores above \tappend{} in 26 of 27 reported combinations
  of model, task, and metric.
\end{itemize}

\section{Related Work}

\paragraph{Long-context architectures and effective use.}
Sparse, compressed, and hybrid sequence models have extended the nominal
context windows of language models
\cite{deepseekai2025deepseekv32,deepseekai2026deepseekv4,
qwen2025qwen3next,yang2025gateddelta}.
Long-context evaluations nevertheless show continued sensitivity to evidence
position, distractors, and the operations required to combine information
across an input \cite{liu2024lostmiddle,needles,ruler}.

\paragraph{Causal order in reasoning.}
Under causal attention, a representation at one position can use only
information from that position and its prefix.
Ok and Lee attribute a large multi-choice prompt-order gap to this
constraint and show that repeating the options after the context partially
closes the gap \cite{ok2026lostpromptorder}.
CoRe reduces sensitivity to the order of supporting documents by repeating the
full context, whereas \emph{Racing Thoughts} traces contextualization errors to
layerwise race conditions \cite{yu2025core,lepori2025racing}.
Collectively, these studies show that input order, repetition, and
contextualization can affect model behavior.

\paragraph{Architectural recurrence.}
Iterative language models can revisit hidden representations or partially
specified text.
Geiping et al. train a language model with recurrent depth that repeatedly
applies a shared block at test time \cite{geiping2025scaling}.
Saunshi et al. study Transformer blocks with shared weights, and
\emph{LoopFormer} trains variable loop trajectories for different inference
budgets \cite{saunshi2025latent,jeddi2026loopformer}.
Masked text-diffusion models such as LLaDA reconstruct masked positions over
multiple steps using visible context on both sides
\cite{sahoo2024maskeddiffusion,nie2025llada}.
These approaches expand the design space for iterative computation. However, they can require substantial changes to existing training and inference infrastructure. 

\paragraph{Rereading and textual feedback.}
Several methods revisit task information or carry textual state forward during
inference.
Re2 repeats the question within one prompt and provides a direct rereading
baseline without prior reasoning text \cite{xu2024rereading}.
The Markovian Thinker carries a bounded text history across reset reasoning
chunks, while ReContext recursively builds and replays an evidence pool for the
current query \cite{markovianthinker,zhao2026recontext}.
These methods show that feeding task-relevant text back to a model can improve
reasoning.

\paragraph{Reasoning traces as state.}
Reasoning traces can serve not only as explanations, but also as textual
records of an evolving computational state.
\cite{hao2026reasoningtraces} shows that synthetic hints inserted into a reasoning trace can affect
later outputs even when follow-up explanations do not acknowledge their
influence.
\emph{State over Tokens} describes the growing reasoning prefix as
externalized computational state, whereas causal mediation evidence suggests
that models do not reliably use their stated intermediate steps
\cite{stateovertokens,paul2024reasoningmatter}.
Taken together, the evidence shows that reasoning traces can carry
task-relevant information and affect later outputs.

\section{Methodology}
\label{sec:method}
Textual reasoning typically follows a causal order, which is well matched
by causal transformers. Yet some reasoning depends on task states whose
values become known only later, creating a mismatch especially
consequential in long contexts. We formalize this mismatch using causal state
update systems and conditional state update tasks and introduce \tas{}.
\begin{table}[t]
\centering
\small
\caption{Vital Notations.}
\label{tab:notations}
\setlength{\tabcolsep}{3pt}
\renewcommand{\arraystretch}{1.06}
\begin{tabular}{@{}p{0.2\columnwidth}p{0.75\columnwidth}@{}}
\toprule
Symbol & Meaning \\
\midrule
$c_i, C$ & The $i$th information unit and the ordered sequence
$C=(c_1,\ldots,c_n)$. \\
$s_i,\mathcal S, b$ & The task state after $c_i$, its finite state space, and the log size of the state space\\
$U$ & The causal state update rule. \\
$z$ & The condition used as the initial task state $s_0$. \\
$x,\mathcal M$ & The long context and the causal reasoning model. \\
$r_j,a_j,\ntr$ & The reasoning trace, visible answer, and number of source model runs. \\
$\pi,T$ & The trace serializer and the resulting serialized trace. \\
\bottomrule
\end{tabular}
\end{table}

 Table~\ref{tab:notations}
summarizes the vital notation used in this section.

\subsection{Causal State Updates and Order-Dependent Memory}
\label{sec:memory}

\paragraph{Causal state updates.}
We define a causal state update processor as a processor that reads each input
unit once in its presented order and updates its persistent working memory using
only its current memory and the newly received unit.

Let $C=(c_1,\ldots,c_n)$ be an ordered sequence of $n$ information units, with
$c_i$ denoting the $i$th unit. As shown in
Figure~\ref{fig:trace-as-state-overview}A, a
causal state update processor carries a task state $s_i$ in a finite state
space $\mathcal S$. After reading $c_i$, it applies a fixed update rule $U$:
\begin{equation}
  s_i=U(s_{i-1},c_i),
  \qquad i=1,\ldots,n.
  \label{eq:update-chain}
\end{equation}
The state $s_i$ is the task state the processor reached after processing
$c_1,\ldots,c_i$. When the initial state $s_0$ is fixed, the processor follows
one realized state path through the sequence.

\paragraph{Conditional state update tasks and order-dependent memory.}
Consider a variant of the causal state update process in which the initial state is supplied by a task condition rather than
being fixed. Let $z\in\mathcal S$ denote this condition. We call the resulting
problem a conditional state update task: the processor sets $s_0=z$, applies
Equation~\ref{eq:update-chain} to $C$, and returns $s_n$.

Figure~\ref{fig:trace-as-state-overview}B compares two orders for the same condition and information sequence.
In the condition first order $[z,C]$, the processor receives $z$ before
$c_1,\ldots,c_n$($n$ is the length of the information sequence) and therefore knows which state path to update. In the
condition last order $[C,z]$, it reads the complete sequence before learning
which initial state should be propagated through it.

To compare their working memory, consider a causal state update processor that knows exactly the state update rule $U$. With $[z,C]$, the processor only needs to
store the current $s_i$, so
$\lceil b\rceil$ bits suffice, where $b=\log_2|\mathcal S|$.

With $[C,z]$, the processor has not selected a state path when it finishes
reading $C$. Each possible sequence determines a complete response profile:
for every $z\in\mathcal S$, the profile specifies the resulting $s_n$. The processor must retain a different configuration for
every distinct response profile induced by the valid sequences.

There are $|\mathcal S|^{|\mathcal S|}$ possible functions from
$\mathcal S$ to itself. In the
worst case, $[C,z]$ requires at least
\begin{equation}
  \left\lceil
    \log_2 |\mathcal S|^{|\mathcal S|}
  \right\rceil
  =\left\lceil
    |\mathcal S|\log_2|\mathcal S|
  \right\rceil=\lceil b 2^b \rceil
  \label{eq:worst-case-order-gap}
\end{equation}
bits, whereas $[z,C]$ requires only
$\lceil b\rceil$ bits. In the worst-case scenario, the memory requirement is exponentially larger in the condition last setting than in the condition first setting. 

We could derive an ordering principle: a task condition could be much easier
to use when it is available before the information whose processing it guides.

\subsection{Reasoning Traces as a Textual State Proxy}
\label{sec:reasoning-traces-carry-state}

Although transformers may retain per-token kv caches, causal transformers with finite context length and finite precision are causal state update models as the their max memory are bounded.  
Conditional state update tasks are not literal models of real world long context tasks. Instead, the results in section~\ref{sec:memory} motivate us to test placing task relevant state before the context on a later pass.

The formal condition $z$ represents task state that is already available to
the processor. In a long context reasoning problem, however, useful task state may
be discovered only while the model is reasoning after reading the context. A
reasoning trace may record useful task states like an active target, a resolved reference or a search
frontier.  

We view reasoning traces as an observable textual proxy for task state, as illustrated in figure~\ref{fig:trace-as-state-overview}C. The
proxy may be incomplete, lossy, or incorrect. We therefore do not identify a
trace with the formal condition $z$ or with a privileged internal model state. 
The connection is functional: if the trace contains useful state information,
placing it before the context can make that information available while the
context is processed again.

Let $\mathcal M$ be a causal state update model and let $x$ denote
the long context. We run $\mathcal M$ $\ntr$ times on the same problem. 
Each run $j$ produces a reasoning trace $r_j$ and a separate visible answer
$a_j$:
\begin{equation}
  (r_j,a_j)\sim\mathcal M(\cdot\mid x),
  \qquad j=1,\ldots,\ntr.
  \label{eq:source-reasoning}
\end{equation}

For each task, a serializer $\pi$ held fixed across the placement
conditions constructs the serialized trace
\begin{equation}
  T=\pi(r_1,\ldots,r_{\ntr}).
  \label{eq:serialized-trace}
\end{equation}
The serializer preserves the included reasoning text in source order and adds
fixed labels and delimiters. We use $T$ as the textual state proxy to be tested. 

\subsection{Trace as state.}
\label{sec:method-tas}

Figure~\ref{fig:trace-as-state-overview}D shows the complete
procedure. Based on theoretical analysis from section~\ref{sec:memory}, we introduce \tas{}, a method that places the textual state proxy $T$ before the long context $x$. 
Therefore, in \tas{}, a fresh causal pass receives $[T,x]$ to generate the answer again. We compare \tas{} with \tappend{}, the method that maintains the original order of $x$ and $T$ and sends the model $[x,T]$ on a second pass. 

Let $r'$ and $a'$ denote the reasoning trace and visible answer produced in this
pass:
\begin{equation}
  \begin{aligned}
    (r'_{\mathrm{Tap}},a'_{\mathrm{Tap}})
      &\sim \mathcal M(\cdot\mid[x,T]),\\
    (r'_{\mathrm{Tas}},a'_{\mathrm{Tas}})
      &\sim \mathcal M(\cdot\mid[T,x]).
  \end{aligned}
  \label{eq:matched-placements}
\end{equation}
\tas{} and \tappend{} use the same long context $x$ and the same textual task state proxy $T$, with order as the only difference.
With \tas{}, $T$ is available while the model processes $x$ again. With
\tappend{}, $T$ arrives after $x$ and cannot change representations already
formed for the preceding context tokens, although it can still influence
subsequent reasoning and the visible answer. The
comparison therefore tests whether the same $T$ is more useful
during rereading than after the long context has already been processed.

\section{Experiments}
\label{sec:experiments}

\subsection{Setup}
\label{sec:experiments-setup}

In this section, we evaluate 
\tas{} and its matched placement control \tappend{} on long context tasks from
different domains. The models come from different providers and have different disclosed
architectural designs. A fixed serializer preserves the reasoning traces while
adding only necessary delimiters such as
"<trace\_start>" and "<trace\_end>" and brief introductory text to construct
$T$. Some traces are so long that we truncate them to the first 50,000
characters to keep the second-pass prompt within the model's context capacity.
These choices define the realization evaluated here; the general \tas{}
framework also permits adapted models, serializers, or state interfaces while
retaining causal processing within each pass.

\paragraph{Models.}
We evaluate Qwen 3.7 Max, DeepSeek V4 Pro Preview, and GLM-5.2. All three are
frontier long-context reasoning models that expose reasoning traces that can be
logged and supplied back to the model as an imperfect textual proxy that may
carry task-state information. They also represent
different providers and disclosed long-context designs: DeepSeek V4 Pro Preview uses a hybrid attention
stack with Compressed Sparse Attention(CSA), Heavily Compressed Attention(HCA), and
Manifold-Constrained Hyper-Connections(mHC) \cite{deepseekai2026deepseekv4};
Qwen 3.7 Max is formed with Gated Deltanet(GDN) and Gated Attention(GA)~\cite{yang2025gateddelta,qiu2025gatedattention}; and GLM-5.2 is a 1M-token
MoE model with DeepSeek Sparse Attention (DSA) optimized by IndexCache (IC),
which reuses sparse-attention indices across layers
\cite{deepseekai2025deepseekv32,bai2026indexcache,glm5team2026glm5,
zaiorg2026glm5repo,zai2026glm52docs}. We choose the
highest available reasoning effort in our experiments: \texttt{max} for
DeepSeek V4 Pro and GLM-5.2. For Qwen 3.7 Max, we use the official  \texttt{xhigh} system prompt. All three support approximately one-million-token inputs in the evaluated
interfaces and expose the reasoning traces required by the frozen textual
realization evaluated here
\cite{alibabacloud2026qwen37max,deepseek2026v4pricing,zai2026glm52docs}. 

Details of the models are described in Table~\ref{tab:model-setup}.

\paragraph{Datasets.}
We use GraphWalks \cite{openai2025graphwalks}, MRCRv2 8-needle
\cite{openai2025mrcr,vodrahalli2024michelangelo}, and 1M-Novel Understanding
Bench (NUB-1M)~\footnote{\url{https://novel-understanding.top}} for our
evaluation.
GraphWalks requires the model to maintain graph state over a long edge list;
MRCRv2 asks the model to bind a final request to the correct earlier
request-response instance. Both provide multiple prompt-length bins. As
different models use different tokenizers, many problems in the 1M bin cannot be fairly tested
due to overlength. We therefore choose the longest bins under 1M: The 256K bin for GraphWalks and the 256K
and 512K bins for MRCRv2. This choice also leaves
room for additional \tas{} and \tappend{} texts.

NUB-1M is a long novel reading comprehension benchmark with complex questions about a new novel containing 400–700K tokens. The dataset is updated by season to reduce leakage
risk. For each season, the problems and answers are manually maintained. 
We use the season 2 novel for evaluation and report average
accuracy of the 20 problems over 5 repeats.

Models may not behave as intended unless the question appears at the end of the prompt.  
We therefore separate the question from the long context and place it at the end of every input to ensure models behave well focused on the given tasks. The literal orders are therefore
\tas{} $[T,x,q]$ and \tappend{} $[x,T,q]$, where $x$ is the long context and
$q$ is the question. 

Details of the datasets are listed in Table~\ref{tab:dataset-setup}.

For each model and task, we compare a single-pass baseline $\mathcal M([x,q])$ with two trace-backed
conditions \tas{}, $\mathcal M([T,x,q])$ and \tappend{} $\mathcal M([x,T,q])$ that reuse the model's own first-pass reasoning traces.
 Appendix~\ref{app:prompt-templates} gives
the trace serialization and dataset-specific insertion boundaries.

\providecommand{\theHtable}{\thetable}

\begingroup
\renewcommand{\thetable}{1-1}
\renewcommand{\theHtable}{setup.1}
\begin{table}[t]
  \centering
  \setlength{\tabcolsep}{2pt}
  \renewcommand{\arraystretch}{1.08}
  \caption{Model configuration and token budgets used in the experiments.}
  \label{tab:model-setup}
  \begin{threeparttable}
  \begin{tabular}{@{}>{\raggedright\arraybackslash}p{0.22\columnwidth}
                  >{\raggedright\arraybackslash}p{0.23\columnwidth}
                  >{\raggedright\arraybackslash}p{0.24\columnwidth}
                  >{\raggedright\arraybackslash}p{0.20\columnwidth}@{}}
    \toprule
  Model & \shortstack[l]{Qwen\\3.7 Max} &
  \shortstack[l]{DeepSeek V4\\Pro Preview} &
  GLM-5.2 \\
    \midrule
    Arch.\tnote{a} &
    \shortstack[l]{GDN+GA} &
    \shortstack[l]{CSA+HCA\\+ mHC} &
    DSA+IC \\
    Input Budget &
    983,616 &
    1,048,576 &
    1,048,576 \\
    Output Budget &
    65,536 &
    131,072 &
    65,536 \\
    Reasoning &
    \texttt{xhigh}\tnote{b} &
    \texttt{max} &
    \texttt{max} \\
    \bottomrule
  \end{tabular}
  \begin{tablenotes}[para,flushleft]
    \footnotesize\raggedright
	 \item[a] Architecture.
    \item[b] Via \texttt{xhigh} prompt.
  \end{tablenotes}
  \end{threeparttable}
\vspace{-0.1in}
\end{table}
\endgroup

\begingroup
\renewcommand{\thetable}{1-2}
\renewcommand{\theHtable}{setup.2}
\begin{table}[t]
  \centering
  \setlength{\tabcolsep}{2pt}
  \renewcommand{\arraystretch}{1.08}
  \caption{Datasets and scoring used in the experiments.}
  \label{tab:dataset-setup}
  \begin{threeparttable}
  \begin{tabular}{@{}>{\raggedright\arraybackslash}p{0.20\columnwidth}
                  >{\raggedright\arraybackslash}p{0.25\columnwidth}
                  >{\raggedright\arraybackslash}p{0.25\columnwidth}
                  >{\raggedright\arraybackslash}p{0.20\columnwidth}@{}}
    \toprule
    Task & GraphWalks & MRCRv2 &
    NUB-1M \\
    \midrule
    Subsets &
    256K &
    \shortstack[l]{256K,512K\\8-needle} &
    Season 2 \\
    Problems &
    200 &
    200 &
    20 \\
    Repeats &
    5 &
    5 &
    5 \\
    Scoring &
    EM\tnote{a}, set F1 &
    EM, Seq.\tnote{b} &
    Acc.\tnote{c} \\
    \bottomrule
  \end{tabular}
  \begin{tablenotes}[para,flushleft]
    \footnotesize
    \item[a] Exact Match. 
    \item[b] SequenceMatcher ratio.
    \item[c] Accuracy.
  \end{tablenotes}
  \end{threeparttable}
\end{table}
\endgroup
\setcounter{table}{1}

We evaluate each problem with 5 repeats and include all 5 reasoning traces in
the second-pass \tas{} and \tappend{} prompts. We use the official provider for these models, 
Aliyun Bailian for Qwen 3.7 Max~\cite{alibabacloud2026qwen37max},
DeepSeek for DeepSeek V4 Pro Preview~\cite{deepseek2026v4pricing}, and
Bigmodel for GLM-5.2~\cite{zai2026glm52docs}. 
We do not explicitly pass a custom maximum-output value. The run records
therefore establish that no client-side cap was requested, but not the
effective provider default, which may also change over time.

Our evaluations use EM and set F1 for GraphWalks, EM and SequenceMatcher ratio
for MRCRv2 8-needle, and DeepSeek V4 Pro model-judged accuracy for NUB-1M.
 Blocked, overlong, malformed, missing, nonterminal, or content-filtered cases
are scored as failures.

Some first pass reasoning traces are very long, we therefore truncate each reused trace to its first 50{,}000
characters to avoid input overlength.

\subsection{Main Results}
\label{sec:experiments-results}

\begin{table*}[!t]
  \centering
  \normalsize
  \setlength{\tabcolsep}{3pt}
  \renewcommand{\arraystretch}{1.12}
  \caption{Long-context results for first pass, \tas{} and \tappend{}. Scores
  are percentages averaged over 5 repeats; higher is better. The best score
  within each model-metric column is \textbf{bolded}.}
  \label{tab:main-results}
  \begin{threeparttable}
  \begin{tabular}{llccccccccc}
    \toprule
    \multirow{3}{*}{Model} &
    \multirow{3}{*}{Condition} &
    \multicolumn{4}{c}{GraphWalks 256K} &
    \multicolumn{4}{c}{MRCRv2 8-needle} &
    NUB-1M \\
    \cmidrule(lr){3-6}\cmidrule(lr){7-10}
    & &
    \multicolumn{2}{c}{BFS} &
    \multicolumn{2}{c}{Parents} &
    \multicolumn{2}{c}{256K} &
    \multicolumn{2}{c}{512K} & Season 2\\
    \cmidrule(lr){3-4}\cmidrule(lr){5-6}
    \cmidrule(lr){7-8}\cmidrule(lr){9-10}
    & & EM & F1 & EM & F1 & EM & Seq. &
    EM & Seq. & Acc. \\
    \midrule
    \multirow{3}{*}{\shortstack[l]{DeepSeek V4\\Pro Preview}}
      & First Pass & 31.6 & 36.4 & 29.2 & 46.5 & 53.8 & 78.5 & 45.4 & 63.6 & 60.0 \\
      & \tappend & 41.8 & 48.3 & 43.0 & 65.3 & 66.6 & 79.6 & 49.4 & 66.9 & 71.0 \\
      & \tas & \textbf{58.8} & \textbf{65.9} & \textbf{81.8} & \textbf{91.3} &
        \textbf{76.2} & \textbf{88.5} & \textbf{52.6} & \textbf{73.3} & \textbf{73.0}\\
    \midrule
    \multirow{3}{*}{\shortstack[l]{Qwen 3.7\\Max}}
      & First Pass & 60.0 & 68.1 & 60.8 & 87.2 & 79.8 & 83.1 & 36.2 & 43.7 & 37.0 \\
      & \tappend & 60.4 & 70.4 & 71.0 & 91.7 & 84.0 & 87.1 & 40.4 & 47.4 & 49.0 \\
      & \tas & \textbf{63.8} & \textbf{71.7} & \textbf{96.4} & \textbf{99.1} &
        \textbf{88.4} & \textbf{91.3} & \textbf{46.8} & \textbf{52.5} & \textbf{51.0} \\
    \midrule
    \multirow{3}{*}{GLM-5.2}
      & First Pass & 55.8 & 70.7 & 66.4 & 88.5 & 40.2 & 48.2 & 42.6 & 52.0 & 43.0 \\
      & \tappend & 60.0 & \textbf{75.8} & 83.2 & 92.7 & 40.0 & 58.3 & 44.6 & 59.9 & 65.0 \\
      & \tas & \textbf{63.4} & 75.0 & \textbf{100.0} & \textbf{100.0} &
        \textbf{61.2} & \textbf{71.9} & \textbf{55.4} & \textbf{70.1} & \textbf{66.0} \\
    \bottomrule
  \end{tabular}
  \end{threeparttable}
\end{table*}

Table~\ref{tab:main-results} displays the main experimental results.
Across all 27 reported combinations of model, task, and metric, \tas{} scores
higher than \tappend{} in 26. DeepSeek V4 Pro and Qwen 3.7 Max favor \tas{} on every reported dataset, task and 
metric. GLM-5.2 follows the same pattern except on GraphWalks BFS F1, where
\tappend{} is 0.8 points higher while \tas{} has higher exact match. \tas{}
also scores above the first pass in all reported evaluations.

\tappend{} improves over the first pass in many settings, showing that
the trace text can carry useful information in these settings. The additional
advantage of \tas{} is consistent with the hypothesis that placing reasoning traces as an imperfect textual task state proxy
before the context is beneficial when the model processes the context again.

The strongest gains occur on GraphWalks Parents, where success
requires the model to maintain predecessor state while interpreting the graph.
DeepSeek V4 Pro improves from 46.5 to 91.3 F1, or 29.2 to 81.8 EM, and Qwen
3.7 Max improves from 87.2 to 99.1 F1, or 60.8 to 96.4 EM. GLM-5.2 reaches
100.0 EM and F1 on Parents under \tas{}. 
MRCRv2 retrieval and binding comparisons show the same ordering
advantage. NUB-1M runs provide supporting evidence in
the same direction for detailed long-novel reading comprehension. 

\subsection{Context Order Ablations}
\label{sec:experiments-controls}

We next compare \tas{} with controls that vary how different context are placed and first-pass outputs are
reused on DeepSeek V4 Pro GraphWalks 256K in table~\ref{tab:graphwalk-ablations}.
We test several ablations to examine how models behave under different trace
placements and feedback controls. For each ablation, the table labels the
prompt format and reports the exact match(EM) and F1 score.
We also include Majority@5 and Oracle@5 for the first pass in the table. Majority@5 evaluates the major choices of the 5 first pass answers,
while Oracle@5 evaluates the best answer among the 5 answers. 

\begin{table*}[!t]
  \centering
  \setlength{\tabcolsep}{2pt}
  \renewcommand{\arraystretch}{1.08}
  \caption{DeepSeek V4 Pro Preview performance on GraphWalks 256K under
  different prompt conditions. Scores are percentages averaged over 5 repeats.
  The best scores for each subtask/metric are
  \textbf{bolded}.}
  \label{tab:graphwalk-ablations}
  \begin{threeparttable}
  \begin{tabular}{@{}lccccc@{}}
    \toprule
    \multirow{2}{*}{Condition} & \multirow{2}{*}{Prompt} &
    \multicolumn{2}{c}{BFS} & \multicolumn{2}{c}{Parents}  \\
    \cmidrule(lr){3-6}
    & & Exact Match(\%) & F1 Score(\%) & Exact Match(\%) & F1 Score(\%)  \\
    \midrule
    First pass & $[x,q]$ & 31.6 & 36.4 & 29.2 & 46.5  \\
    Majority@5 &  & 35.0 & 39.6 & 31.0 & 47.8 \\
    Oracle@5 &  & 50.0 & 55.5 & 50.0 & 75.0  \\
    Question First & $[q,x,q]$ & 33.4 & 36.8 & 53.0 & 64.6 \\
    Re2~\cite{xu2024rereading} & $[x,q,x,q]$ & 49.0 & 52.9 & 50.0 & 68.9  \\
    Answer Feedback & $[a,x,q]$ & 35.4 & 39.4 & 45.4 & 58.0  \\
    Random Trace & $[T_{\mathrm{rand}},x,q]$ & 22.4 & 35.9 & 14.2 & 31.3  \\
    Trace Only & $[T,q]$ & 46.2 & 52.7 & 43.8 & 67.3  \\
    \midrule
    \tappend{} & $[x,T,q]$ & 41.8 & 48.3 & 43.0 & 65.3  \\
    \tas{} & $[T,x,q]$ & \textbf{58.8} & \textbf{65.9} &
    \textbf{81.8} & \textbf{91.3} \\
    \bottomrule
  \end{tabular}
  \end{threeparttable}
\end{table*}
Question First puts the question $q$ before the task. It substantially improves performance on Parents but remains similar to the first pass baseline on BFS, implying that the question itself may be a vital task state for some tasks. 

Re2~\cite{xu2024rereading} improves substantially over the first pass, showing that rereading is useful on these long context tasks. However, it remains below \tas{},
especially on Parents. In this evaluated setting, the comparison is consistent
with $T$ carrying task-relevant information beyond that supplied by prompt
repetition alone.

Answer Feedback places the first pass answers $a=[a_1,\ldots,a_{\ntr}]$ before
the original prompt. It improves over the first pass but remains far below
\tas{}. This supports
the interpretation that serialized reasoning text can serve
as a more useful task state proxy than the answers or the questions alone.

Random Trace replaces the prefix with traces sampled from other problems in the
same GraphWalks subtask. It performs worse than the no-trace first pass, much worse than \tas{}.
So the gain of \tas{} is not due to a generic formatting effect, and a
reasoning-trace-like scaffold alone is insufficient without problem-specific,
task-relevant information from reasoning traces of the same problem.

All above controls above place a copy of the long context late in the prompt
but remain below \tas{}, signifying that textual recency alone does not explain the observed \tas{} gains.

Trace Only sends the first-pass reasoning traces without the original long
context prompt. Its performance is similar to 
\tappend{}, showing that the traces contain useful answer-relevant information.
Nevertheless, it remains well below \tas{}, showing that the original input is still useful when placed 
after the first pass traces. 
Trace as State also outperforms Oracle@5, showing that the feedback pass improves on what can be obtained by retrospectively selecting the best of the five first pass outputs.

\subsection{Trace Count Ablation}
\label{sec:experiments-ablation}

Finally, we rerun DeepSeek V4 Pro Preview on the GraphWalks 256K
while varying the number of reused first-pass traces from $\ntr=1$ to
$\ntr=5$.  The block for each count contains the first $\ntr$ eligible traces
in repeat order, so successive settings are nested.  For every $\ntr\geq1$,
\tas{} and \tappend{} receive the same realized $T$, and each condition
averages five fresh second-pass repeats per problem.  The $\ntr=0$ point is the common
five-repeat first-pass mean.
\begin{figure}[!t]
  \centering
  \includegraphics[width=\columnwidth]{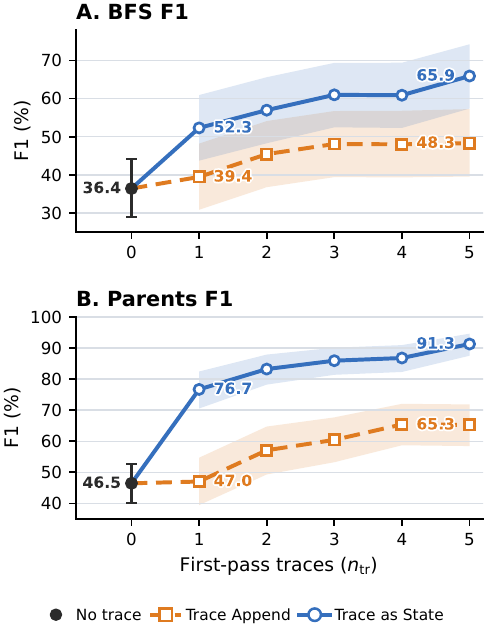}
  \caption{Trace-count ablation on DeepSeek V4 Pro GraphWalks 256K.  Panel A
  reports BFS F1 and Panel B reports Parents F1. Shading shows 95\% percentile
  confidence intervals.}
  \label{fig:trace-count-ablation}
\end{figure}
Figure~\ref{fig:trace-count-ablation} shows that performance
generally rises as additional traces are included, and \tas{} remains above
\tappend{} for every $\ntr\geq1$.
Within this frozen-model realization, the result supports trace count as an
inference-scaling parameter and shows that the measured \tas{}--\tappend{}
ordering persists from $\ntr=1$ through $\ntr=5$.

\section{Conclusion}

We introduce \tas{}, an inference approach that reuses task state information
carried in reasoning traces as a textual proxy on a fresh pass over a
long context task. Its motivation comes from theoretical analysis of
conditional state update tasks: for causal state update processors, a condition available before an information sequence can guide a
single evolving state, while a late condition can require retaining much more memory about potential conditions of the sequence.

\tas{} applies this ordering principle by making prior reasoning available
before the long context block is processed again. Together, the formal analysis and
experiments support a simple view: reasoning traces can carry forward task
state information, and the point at which that information becomes available
can shape how useful it is.

This view suggests several direct extensions: selecting or compressing traces,
learning better textual state interfaces, and optimizing where feedback is
placed. A broader training framework could jointly learn the model and the feedback interface while retaining causal processing within each pass.

\tas{} uses one or more source runs followed by a fresh pass over the task.
These additional passes increase inference latency and token cost. Placing
state before the original context can also reduce key-value cache reuse in
multi-round agentic settings. Another direction is to find more efficient ways that maintain the benefits of \tas{} while reducing the scale of additional costs.

\clearpage
\section*{Limitations}

Our experiments realize cross-pass state feedback through model-generated
text. This evaluated realization requires access to raw reasoning traces or
another exposed state interface. The requirement comes from the interface used
in our experiments. The general cross-pass design can instead use any
accessible state interface. Models or APIs that expose only final answers may
therefore require a different interface.

Our evaluation covers three causal transformer models from different providers
and three long context task families: GraphWalks, MRCR, and NUB-1M. We do not
evaluate multi-round agent tasks. Testing additional models, domains, context
lengths, and interactive settings is needed to establish how broadly the
observed placement advantage generalizes.

Finally, because the method reuses only model generated traces, it does not introduce additional ethical concerns or misuse risks beyond those associated with the underlying models and tasks.

\section*{Ethical Considerations}
\tas{} probes little potential risks in single-turn long context tasks described in this paper, as the model can only output the reasoning contents and an answer and cannot interact with the environment in these tasks.

However, in further research of applying \tas{} on agentic coding setups, we discover an increased rate of attempts to break out of the sandbox under \tas{} than under normal interleaved thinking using the same models, as \tas{} may emphasis some potentially risky reasoning contents of the models. This finding underscores the need for transparent reasoning traces and robust monitoring to detect potentially unsafe behavior in agentic coding systems.

\clearpage
\bibliography{references}

@inproceedings{needles,
  title     = {{BABIL}ong: Testing the Limits of {LLM}s with Long Context Reasoning-in-a-Haystack},
  author    = {Yuri Kuratov and Aydar Bulatov and Petr Anokhin and Ivan Rodkin and Dmitry Sorokin and Artyom Sorokin and Mikhail Burtsev},
  booktitle = {Advances in Neural Information Processing Systems},
  volume    = {37},
  year      = {2024},
  doi       = {10.52202/079017-3381},
  url       = {https://papers.nips.cc/paper_files/paper/2024/hash/c0d62e70dbc659cc9bd44cbcf1cb652f-Abstract-Datasets_and_Benchmarks_Track.html},
  note      = {Datasets and Benchmarks Track}
}

@inproceedings{saunshi2025latent,
  title     = {Reasoning with Latent Thoughts: On the Power of Looped Transformers},
  author    = {Nikunj Saunshi and Nishanth Dikkala and Zhiyuan Li and Sanjiv Kumar and Sashank J. Reddi},
  booktitle = {The Thirteenth International Conference on Learning Representations},
  year      = {2025}
}

@inproceedings{geiping2025scaling,
  title     = {Scaling up Test-Time Compute with Latent Reasoning: A Recurrent Depth Approach},
  author    = {Jonas Geiping and Sean Michael McLeish and Neel Jain and John Kirchenbauer and Siddharth Singh and Brian R. Bartoldson and Bhavya Kailkhura and Abhinav Bhatele and Tom Goldstein},
  booktitle = {The Thirty-ninth Annual Conference on Neural Information Processing Systems},
  year      = {2025}
}

@inproceedings{jeddi2026loopformer,
  title     = {{LoopFormer}: Elastic-Depth Looped Transformers for Latent Reasoning via Shortcut Modulation},
  author    = {Ahmadreza Jeddi and Marco Ciccone and Babak Taati},
  booktitle = {The Fourteenth International Conference on Learning Representations},
  year      = {2026}
}

@inproceedings{loopformer,
  title     = {{LoopFormer}: Elastic-Depth Looped Transformers for Latent Reasoning via Shortcut Modulation},
  author    = {Ahmadreza Jeddi and Marco Ciccone and Babak Taati},
  booktitle = {The Fourteenth International Conference on Learning Representations},
  year      = {2026}
}

@inproceedings{merrill2024expressive,
  title     = {The Expressive Power of Transformers with Chain of Thought},
  author    = {William Merrill and Ashish Sabharwal},
  booktitle = {The Twelfth International Conference on Learning Representations},
  year      = {2024}
}

@inproceedings{yang2025gateddelta,
  title     = {Gated Delta Networks: Improving {Mamba2} with Delta Rule},
  author    = {Songlin Yang and Jan Kautz and Ali Hatamizadeh},
  booktitle = {The Thirteenth International Conference on Learning Representations},
  year      = {2025}
}

@inproceedings{qiu2025gatedattention,
  title     = {Gated Attention for Large Language Models: Non-Linearity, Sparsity, and Attention-Sink-Free},
  author    = {Zihan Qiu and Zekun Wang and Bo Zheng and Zeyu Huang and Kaiyue Wen and Songlin Yang and Rui Men and Le Yu and Fei Huang and Suozhi Huang and Dayiheng Liu and Jingren Zhou and Junyang Lin},
  booktitle = {Advances in Neural Information Processing Systems},
  volume    = {38},
  pages     = {110931--110957},
  year      = {2025},
  doi       = {10.52202/085713-3345},
  url       = {https://proceedings.neurips.cc/paper_files/paper/2025/hash/904e89bb4e632e75fb47f093b620b257-Abstract-Conference.html}
}

@misc{scratchpad,
  title  = {Show Your Work: Scratchpads for Intermediate Computation with Language Models},
  author = {Maxwell Nye and Anders Johan Andreassen and Guy Gur-Ari and Henryk Michalewski and Jacob Austin and David Bieber and David Dohan and Aitor Lewkowycz and Maarten Bosma and David Luan and Charles Sutton and Augustus Odena},
  year   = {2021},
  note   = {arXiv:2112.00114}
}

@inproceedings{cot,
  title     = {Chain-of-Thought Prompting Elicits Reasoning in Large Language Models},
  author    = {Jason Wei and Xuezhi Wang and Dale Schuurmans and Maarten Bosma and Brian Ichter and Fei Xia and Ed Chi and Quoc V. Le and Denny Zhou},
  booktitle = {Advances in Neural Information Processing Systems},
  volume    = {35},
  pages     = {24824--24837},
  year      = {2022}
}

@inproceedings{paul2024reasoningmatter,
  title     = {Making Reasoning Matter: Measuring and Improving Faithfulness of Chain-of-Thought Reasoning},
  author    = {Debjit Paul and Robert West and Antoine Bosselut and Boi Faltings},
  booktitle = {Findings of the Association for Computational Linguistics: EMNLP 2024},
  pages     = {15012--15032},
  year      = {2024},
  address   = {Miami, Florida, USA},
  publisher = {Association for Computational Linguistics},
  doi       = {10.18653/v1/2024.findings-emnlp.882},
  url       = {https://aclanthology.org/2024.findings-emnlp.882/}
}

@article{deepr1,
  title   = {{DeepSeek-R1}: Incentivizing Reasoning Capability in {LLMs} via Reinforcement Learning},
  author  = {{DeepSeek-AI}},
  journal = {Nature},
  volume  = {645},
  pages   = {633--638},
  year    = {2025},
  doi     = {10.1038/s41586-025-09422-z}
}

@misc{stateovertokens,
  title         = {State over Tokens: Characterizing the Role of Reasoning Tokens},
  author        = {Mosh Levy and Zohar Elyoseph and Shauli Ravfogel and Yoav Goldberg},
  year          = {2025},
  eprint        = {2512.12777},
  archivePrefix = {arXiv},
  primaryClass  = {cs.CL},
  url           = {https://arxiv.org/abs/2512.12777}
}

@inproceedings{hao2026reasoningtraces,
  title     = {Reasoning Traces Shape Outputs but Models Won{'}t Say So},
  author    = {Hao, Yijie and Chen, Lingjie and Emami, Ali and Ho, Joyce C.},
  booktitle = {Proceedings of the 64th Annual Meeting of the Association for Computational Linguistics (Volume 1: Long Papers)},
  month     = jul,
  pages     = {42852--42878},
  year      = {2026},
  address   = {San Diego, California, United States},
  publisher = {Association for Computational Linguistics},
  doi       = {10.18653/v1/2026.acl-long.1986},
  url       = {https://aclanthology.org/2026.acl-long.1986/}
}

@inproceedings{criticalthinking,
  title     = {Critical Thinking: Which Kinds of Complexity Govern Optimal Reasoning Length?},
  author    = {Celine Lee and Alexander M. Rush and Keyon Vafa},
  booktitle = {Proceedings of the 14th International Joint Conference on Natural Language Processing and the 4th Conference of the Asia-Pacific Chapter of the Association for Computational Linguistics},
  year      = {2025},
  doi       = {10.18653/v1/2025.ijcnlp-long.57}
}

@inproceedings{markovianthinker,
  title     = {The Markovian Thinker: Architecture-Agnostic Linear Scaling of Reasoning},
  author    = {Milad Aghajohari and Kamran Chitsaz and Amirhossein Kazemnejad and Sarath Chandar and Alessandro Sordoni and Aaron Courville and Siva Reddy},
  booktitle = {The Fourteenth International Conference on Learning Representations},
  year      = {2026},
  url       = {https://openreview.net/forum?id=3As6AQ9ELI},
  note      = {Poster}
}

@article{liu2024lostmiddle,
  title     = {Lost in the Middle: How Language Models Use Long Contexts},
  author    = {Nelson F. Liu and Kevin Lin and John Hewitt and Ashwin Paranjape and Michele Bevilacqua and Fabio Petroni and Percy Liang},
  journal   = {Transactions of the Association for Computational Linguistics},
  volume    = {12},
  pages     = {157--173},
  year      = {2024},
  publisher = {MIT Press},
  doi       = {10.1162/tacl_a_00638},
  url       = {https://aclanthology.org/2024.tacl-1.9/}
}

@inproceedings{ok2026lostpromptorder,
  title     = {Lost in the Prompt Order: Revealing the Limitations of Causal Attention in Language Models},
  author    = {Hyunjong Ok and Jaeho Lee},
  booktitle = {Findings of the Association for Computational Linguistics: ACL 2026},
  month     = jul,
  year      = {2026},
  address   = {San Diego, California, United States},
  publisher = {Association for Computational Linguistics},
  pages     = {38566--38587},
  doi       = {10.18653/v1/2026.findings-acl.1921},
  url       = {https://aclanthology.org/2026.findings-acl.1921/}
}

@inproceedings{chen2024premiseorder,
  title     = {Premise Order Matters in Reasoning with Large Language Models},
  author    = {Xinyun Chen and Ryan Andrew Chi and Xuezhi Wang and Denny Zhou},
  booktitle = {Proceedings of the 41st International Conference on Machine Learning},
  series    = {Proceedings of Machine Learning Research},
  volume    = {235},
  pages     = {6596--6620},
  month     = jul,
  year      = {2024},
  publisher = {PMLR},
  url       = {https://proceedings.mlr.press/v235/chen24i.html}
}

@inproceedings{xu2024rereading,
  title     = {Re-Reading Improves Reasoning in Large Language Models},
  author    = {Xiaohan Xu and Chongyang Tao and Tao Shen and Can Xu and Hongbo Xu and Guodong Long and Jian-Guang Lou and Shuai Ma},
  booktitle = {Proceedings of the 2024 Conference on Empirical Methods in Natural Language Processing},
  month     = nov,
  pages     = {15549--15575},
  year      = {2024},
  address   = {Miami, Florida, USA},
  publisher = {Association for Computational Linguistics},
  doi       = {10.18653/v1/2024.emnlp-main.871},
  url       = {https://aclanthology.org/2024.emnlp-main.871/}
}

@inproceedings{yu2025core,
  title     = {Unleashing Multi-Hop Reasoning Potential in Large Language Models through Repetition of Misordered Context},
  author    = {Sangwon Yu and Ik-hwan Kim and Jongyoon Song and Saehyung Lee and Junsung Park and Sungroh Yoon},
  booktitle = {Findings of the Association for Computational Linguistics: NAACL 2025},
  month     = apr,
  year      = {2025},
  address   = {Albuquerque, New Mexico},
  publisher = {Association for Computational Linguistics},
  pages     = {6450--6470},
  doi       = {10.18653/v1/2025.findings-naacl.360},
  url       = {https://aclanthology.org/2025.findings-naacl.360/}
}

@misc{zhao2026recontext,
  title         = {{ReContext}: Recursive Evidence Replay as {LLM} Harness for Long-Context Reasoning},
  author        = {Yanjun Zhao and Ruizhong Qiu and Tianxin Wei and Yuanchen Bei and Zhining Liu and Lingjie Chen and Ismini Lourentzou and Hanghang Tong and Jingrui He},
  year          = {2026},
  eprint        = {2607.02509},
  archivePrefix = {arXiv},
  primaryClass  = {cs.AI},
  url           = {https://arxiv.org/abs/2607.02509}
}

@inproceedings{ruler,
  title     = {{RULER}: What's the Real Context Size of Your Long-Context Language Models?},
  author    = {Cheng-Ping Hsieh and Simeng Sun and Samuel Kriman and Shantanu Acharya and Dima Rekesh and Fei Jia and Yang Zhang and Boris Ginsburg},
  booktitle = {First Conference on Language Modeling},
  year      = {2024}
}

@inproceedings{lepori2025racing,
  title     = {Racing Thoughts: Explaining Contextualization Errors in Large Language Models},
  author    = {Michael A. Lepori and Michael Curtis Mozer and Asma Ghandeharioun},
  booktitle = {Proceedings of the 2025 Conference of the Nations of the Americas Chapter of the Association for Computational Linguistics: Human Language Technologies (Volume 1: Long Papers)},
  month     = apr,
  pages     = {3020--3036},
  year      = {2025},
  address   = {Albuquerque, New Mexico},
  publisher = {Association for Computational Linguistics},
  doi       = {10.18653/v1/2025.naacl-long.155},
  url       = {https://aclanthology.org/2025.naacl-long.155/}
}

@techreport{deepseekai2026deepseekv4,
  title         = {{DeepSeek-V4}: Towards Highly Efficient Million-Token Context Intelligence},
  author        = {{DeepSeek-AI}},
  institution   = {DeepSeek-AI},
  type          = {Technical report},
  year          = {2026},
  number        = {arXiv:2606.19348},
  eprint        = {2606.19348},
  archivePrefix = {arXiv},
  primaryClass  = {cs.CL},
  doi           = {10.48550/arXiv.2606.19348},
  url           = {https://arxiv.org/abs/2606.19348}
}

@techreport{deepseekai2025deepseekv32,
  title         = {{DeepSeek-V3.2}: Pushing the Frontier of Open Large Language Models},
  author        = {{DeepSeek-AI}},
  institution   = {DeepSeek-AI},
  type          = {Technical report},
  year          = {2025},
  number        = {arXiv:2512.02556},
  eprint        = {2512.02556},
  archivePrefix = {arXiv},
  primaryClass  = {cs.CL},
  doi           = {10.48550/arXiv.2512.02556},
  url           = {https://arxiv.org/abs/2512.02556}
}

@techreport{qwen2025qwen3technical,
  title         = {{Qwen3} Technical Report},
  author        = {{Qwen Team}},
  institution   = {Alibaba Cloud},
  type          = {Technical report},
  year          = {2025},
  number        = {arXiv:2505.09388},
  eprint        = {2505.09388},
  archivePrefix = {arXiv},
  primaryClass  = {cs.CL},
  doi           = {10.48550/arXiv.2505.09388},
  url           = {https://arxiv.org/abs/2505.09388}
}

@techreport{qwen2025qwen3next,
  title       = {{Qwen3-Next-80B-A3B-Instruct} Model Card},
  author      = {{Qwen Team}},
  institution = {Alibaba Cloud},
  type        = {Model card},
  year        = {2025},
  url         = {https://huggingface.co/Qwen/Qwen3-Next-80B-A3B-Instruct},
  note        = {Official model card}
}

@techreport{glm5team2026glm5,
  title         = {{GLM-5}: From Vibe Coding to Agentic Engineering},
  author        = {{GLM-5 Team}},
  institution   = {Z.ai and Tsinghua University},
  type          = {Technical report},
  year          = {2026},
  number        = {arXiv:2602.15763},
  eprint        = {2602.15763},
  archivePrefix = {arXiv},
  primaryClass  = {cs.LG},
  doi           = {10.48550/arXiv.2602.15763},
  url           = {https://arxiv.org/abs/2602.15763}
}

@inproceedings{bai2026indexcache,
  title     = {{IndexCache}: Accelerating Sparse Attention via Cross-Layer Index Reuse},
  author    = {Bai, Yushi and Dong, Qian and Jiang, Ting and Lv, Xin and Du, Zhengxiao and Zeng, Aohan and Tang, Jie and Li, Juanzi},
  booktitle = {Third Conference on Language Modeling},
  year      = {2026},
  url       = {https://arxiv.org/abs/2603.12201}
}

@misc{zaiorg2026glm5repo,
  title        = {{GLM-5.2} \& {GLM-5.1} \& {GLM-5}},
  author       = {{Z.ai}},
  year         = {2026},
  howpublished = {\url{https://github.com/zai-org/GLM-5}},
  note         = {Official GLM-5 series repository. Accessed August 3, 2026}
}

@misc{zai2026glm52docs,
  title        = {{GLM-5.2}},
  author       = {{Z.ai}},
  year         = {2026},
  howpublished = {\url{https://docs.bigmodel.cn/cn/guide/models/text/glm-5.2}},
  note         = {Official model documentation. Accessed August 3, 2026}
}

@misc{openai2025graphwalks,
  title        = {{GraphWalks}: A Multi Hop Reasoning Long Context Benchmark},
  author       = {{OpenAI}},
  year         = {2025},
  howpublished = {\url{https://huggingface.co/datasets/openai/graphwalks}},
  note         = {Dataset card. Accessed August 3, 2026}
}

@misc{openai2025mrcr,
  title        = {{OpenAI MRCR}: Long Context Multiple Needle in a Haystack Benchmark},
  author       = {{OpenAI}},
  year         = {2025},
  howpublished = {\url{https://huggingface.co/datasets/openai/mrcr}},
  note         = {Dataset card. Accessed August 3, 2026}
}

@inproceedings{sahoo2024maskeddiffusion,
  title     = {Simple and Effective Masked Diffusion Language Models},
  author    = {Sahoo, Subham Sekhar and Arriola, Marianne and Schiff, Yair and Gokaslan, Aaron and Marroquin, Edgar and Chiu, Justin T. and Rush, Alexander and Kuleshov, Volodymyr},
  booktitle = {Advances in Neural Information Processing Systems},
  volume    = {37},
  year      = {2024},
  doi       = {10.52202/079017-4135},
  url       = {https://proceedings.neurips.cc/paper_files/paper/2024/hash/eb0b13cc515724ab8015bc978fdde0ad-Abstract-Conference.html}
}

@inproceedings{nie2025llada,
  title     = {Large Language Diffusion Models},
  author    = {Nie, Shen and Zhu, Fengqi and You, Zebin and Zhang, Xiaolu and Ou, Jingyang and Hu, Jun and Zhou, Jun and Lin, Yankai and Wen, Ji-Rong and Li, Chongxuan},
  booktitle = {Advances in Neural Information Processing Systems},
  volume    = {38},
  pages     = {50608--50646},
  year      = {2025},
  publisher = {Curran Associates, Inc.},
  doi       = {10.52202/085713-1689},
  url       = {https://proceedings.neurips.cc/paper_files/paper/2025/hash/48b383b24230e0e6e649d9c98dae4d8c-Abstract-Conference.html}
}

@misc{vodrahalli2024michelangelo,
  title         = {{Michelangelo}: Long Context Evaluations Beyond Haystacks via Latent Structure Queries},
  author        = {Vodrahalli, Kiran and Ontanon, Santiago and Tripuraneni, Nilesh and Xu, Kelvin and Jain, Sanil and Shivanna, Rakesh and Hui, Jeffrey and Dikkala, Nishanth and Kazemi, Mehran and Fatemi, Bahare and Anil, Rohan and Dyer, Ethan and Shakeri, Siamak and Vij, Roopali and Mehta, Harsh and Ramasesh, Vinay and Le, Quoc and Chi, Ed and Lu, Yifeng and Firat, Orhan and Lazaridou, Angeliki and Lespiau, Jean-Baptiste and Attaluri, Nithya and Olszewska, Kate},
  year          = {2024},
  eprint        = {2409.12640},
  archivePrefix = {arXiv},
  primaryClass  = {cs.CL},
  doi           = {10.48550/arXiv.2409.12640},
  url           = {https://arxiv.org/abs/2409.12640}
}

@misc{deepseek2026v4pricing,
  title        = {Models \& Pricing},
  author       = {{DeepSeek-AI}},
  year         = {2026},
  howpublished = {\url{https://api-docs.deepseek.com/quick_start/pricing/}},
  note         = {Official API documentation. Accessed August 3, 2026}
}

@misc{alibabacloud2026qwen37max,
  title        = {{qwen3.7-max} Model Information},
  author       = {{Alibaba Cloud}},
  year         = {2026},
  howpublished = {\url{https://help.aliyun.com/zh/model-studio/qwen3-7-max}},
  note         = {Official Model Studio documentation. Accessed August 3, 2026}
}
\clearpage
\appendix
\raggedbottom
\makeatletter
\setlength{\@dblfptop}{0pt}
\makeatother
\section{Attaining the Worst-Case Residual-Map Count}
\label{app:worst-case-residual-maps}

Let $\mathcal S$ be a finite nonempty state set, $\mathcal C$ an input
alphabet, $n$ the sequence length, and
$t:\mathcal C^n\times\mathcal S\rightarrow\mathcal S$ a conditional
state-update task. For an information sequence $C\in\mathcal C^n$, define the
residual map and the number of distinct residual maps by
\begin{align*}
  \phi_C
  &=(t(C,z))_{z\in\mathcal S}\in\mathcal S^{\mathcal S},\\
  \mathcal K
  &=\left|\{\phi_C\mid C\in\mathcal C^n\}\right|.
\end{align*}
Equivalently, $\phi_C(z)=t(C,z)$. For every finite nonempty state set
$\mathcal S$, this appendix constructs a task for which
$\mathcal K=|\mathcal S|^{|\mathcal S|}$. The corresponding late-order
working-memory requirement is
\[
  \left\lceil\log_2\mathcal K\right\rceil
  =\left\lceil|\mathcal S|\log_2|\mathcal S|\right\rceil
  \text{ bits}.
\]
Fix an arbitrary finite nonempty state set $\mathcal S$. For this
$\mathcal S$, the construction below chooses $\mathcal C$, $n$, and $t$ and
evaluates $\mathcal K$ under the definition above.
Set $m=|\mathcal S|$ and write
$b=\log_2 m$.  Since every $\phi_C$ belongs to
$\mathcal S^{\mathcal S}$, $\mathcal K\le m^m$.

\paragraph{Proposition.}
For every finite nonempty state set $\mathcal S$, there exist a finite input
alphabet $\mathcal C$ and a conditional state-update task such that, with
$n=1$, $\mathcal K=m^m$.  Every
deterministic one-pass processor that is exact on all inputs and reads $C$
before $z$ must therefore retain at least
\[
  \left\lceil\log_2 \mathcal K\right\rceil
  =
  \left\lceil m\log_2 m\right\rceil
  =
  \left\lceil b\,2^b\right\rceil
\]
bits of persistent input-dependent state immediately before reading $z$.

\paragraph{Construction.}
Set $n=1$ and choose the input alphabet
\[
  \mathcal C=\{c_f\mid f\in\mathcal S^{\mathcal S}\}.
\]
Thus $|\mathcal C|=m^m$.  Define one fixed update rule
$U:\mathcal S\times\mathcal C\to\mathcal S$ by $U(s,c_f)=f(s)$, equivalently
$U_{c_f}=f$.  For every $f\in\mathcal S^{\mathcal S}$, the sequence
$C_f=(c_f)$ belongs to $\mathcal C^n$, and its residual map satisfies
\[
  \phi_{C_f}(z)
  =t(C_f,z)
  =U_{c_f}(z)
  =f(z).
\]
Consequently,
\[
  \{\phi_C\mid C\in\mathcal C^n\}=\mathcal S^{\mathcal S},
\]
and
\begin{align*}
  \mathcal K
  &=\left|\{\phi_C\mid C\in\mathcal C^n\}\right|\\
  &=|\mathcal S^{\mathcal S}|=m^m.
\end{align*}

\paragraph{Late-order cut.}
Consider the processor configuration after $C_f$ and immediately before $z$
is read.  If two distinct functions $f\ne g$ produced the same configuration,
some $z\in\mathcal S$ would satisfy $f(z)\ne g(z)$.  Starting from the shared
configuration, the processor would produce the same output after reading the
identical suffix $z$, contradicting exactness.  Thus the cut admits at least
$m^m$ distinct configurations.  At this cut, a processor attains the bound by
storing the identity of $f$ in one of $m^m$ states and applying the fixed
lookup rule when $z$ arrives.  Under the accounting convention above, the
exact cut-state requirement for this family is
$\lceil\log_2(m^m)\rceil$ bits.

\paragraph{Condition-first comparison.}
For the order $(z,c_f)$, a single $\mathcal S$-valued register suffices:
initialize it with $z$ and replace its value by $f(z)$ when $c_f$ arrives.  At
the cut after $z$ and immediately before $c_f$, this implementation has $m$
possible input-dependent configurations.  This is optimal.  Let
$\iota=\operatorname{id}_{\mathcal S}$, so $c_\iota\in\mathcal C$.  If two
distinct values $z,z'\in\mathcal S$ produced the same configuration at this
cut, then reading the identical suffix $c_\iota$ would force the same output
from both configurations.  Exactness instead requires the respective outputs
$\iota(z)=z$ and $\iota(z')=z'$.  Hence the cut admits at least $m$
configurations, and its exact cut-state requirement is
$\lceil\log_2 m\rceil=\lceil b\rceil$ bits.  When $m=1$, each cut has one
configuration and therefore requires zero bits.  As a concrete check, $m=4$
gives $\mathcal K=4^4=256$: the late-order cut requires eight bits, whereas
the condition-first cut requires two bits.

\paragraph{Scope.}
This finite-state existence construction applies to the stated task family
and to deterministic, exact, one-pass computation.  It is an adversarial
worst-case construction whose alphabet and fixed transition table realize all
self-maps of $\mathcal S$.  The processor cannot reread $C$, and every
auxiliary writable store is included in the counted configuration.  Restricted
update families may realize fewer residual maps, yielding a smaller lower
bound from this residual-map argument.  

\paragraph{Causal Transformers are Causal State Update Processors.}
Transformers use kv caches that may expand as context length grows.
 Modern transformers may include more complex memory structures like latent kv, shared kv or linear kv. Despite these, causal transformers with a finite maximal context length and finite precisions are indeed causal state update processors defined in section~\ref{sec:memory}. 

Define the "working state" as to 
contain the current position, all layerwise key--value entries, and any other persistent input-dependent inference buffers.
Causal inference computes the new token
representation layer by layer and appends the corresponding key--value entries.
Thus the next configuration is a fixed function of the preceding configuration
and the new token. The maximum context length and finite precision make the
configuration space finite. Per-token or other forms of caches therefore do not violate the causal
state update abstraction. 

\section{Scoring and Run Qualifications}
\label{app:experimental-details}

Section~\ref{sec:experiments-setup} gives the common model, dataset, condition,
and scoring setup. This appendix specifies scorer edge cases, qualifications of
the reported runs, and the record-selection rules used to compute the reported
cells.

\subsection{Scorer Edge Cases}

\paragraph{GraphWalks.}
The evaluator extracts the terminal line \texttt{Final Answer: [...]} and
compares the parsed node set with the gold set. A valid empty prediction is
scored normally: two empty sets have EM and F1 equal to one, whereas one empty
and one nonempty set have EM and F1 equal to zero. An absent response or a
missing, malformed, or nonterminal answer receives zero.
For Question First, we conservatively treat a null or unknown termination
marker as length-limited. In the evaluated records, 121 have a null marker,
none have an unknown marker, and 60 are natively marked as length-limited, for
181 length-limited outputs in total (160 BFS and 21 Parents). One additional
stopped BFS output lacks the required terminal answer syntax. All 182 outputs
receive zero.

\paragraph{MRCRv2 8-needle.}
Each example supplies a random prefix that must begin both the prediction and
reference. The scorer validates and removes this prefix before computing EM
and the ratio returned by Python's \texttt{difflib.SequenceMatcher}. EM is one
when the remaining strings match exactly and zero otherwise. The paper labels
the ratio \emph{Seq.}; it is distinct from GraphWalks set F1. A missing or
invalid prefix gives zero for both metrics. Some prompts trigger provider
content filters; filtered outputs remain in the denominator and receive zero.

\paragraph{NUB-1M.}
We maintain a reference answer for each problem and use DeepSeek V4 Pro to
compare the extracted solver answer with that reference. Solver identity and
feedback order are excluded from the judge prompt. The judge returns a
Boolean correctness decision and a short rationale; we use the Boolean as the
binary score. 

We also manually reviewed the judged outputs and found no errors.

\section{Prompt and State Templates}
\label{app:prompt-templates}

This appendix shows the fixed serializer used to construct $T$ and records the
dataset-specific prompt details for experiments.  We retain the compact condition notation $[T,x]$ for \tas{} and
$[x,T]$ for \tappend{}; the paragraphs below give the exact insertion
boundaries and fixed interface text suppressed by that notation.

\subsection{Reusable Trace Block}

Let $r=(r_1,\ldots,r_{\ntr})$ denote the selected first-pass reasoning traces
in source order, and let $P_{\mathcal D}$ denote the dataset-specific
trace-block preamble containing its description and warning.
Table~\ref{tab:trace-serializer} shows their fixed text form.

\begin{table*}[t]
  \centering
  \small
  \setlength{\tabcolsep}{8pt}
  \renewcommand{\arraystretch}{1.08}
  \caption{The fixed serialization used to construct the reusable trace block.}
  \label{tab:trace-serializer}
  \begin{tabular}{@{}>{\raggedright\arraybackslash}p{0.39\textwidth}
                  >{\raggedright\arraybackslash}p{0.55\textwidth}@{}}
    \toprule
    Reasoning traces $r$ & $T=\pi(r)$ \\
    \midrule
    $\displaystyle r=(r_1,r_2,\ldots,r_{\ntr})$
    &
    $P_{\mathcal D}$ \quad \emph{(dataset-specific prompt)}\par
    \texttt{<first\_run\_reasoning\_traces>}\par
    \texttt{[Trace 1]}\par
    $r_1$\par\smallskip
    \texttt{[Trace 2]}\par
    $r_2$\par\smallskip
    $\vdots$\par\smallskip
    \texttt{[Trace }$\ntr$\texttt{]}\par
    $r_{\ntr}$\par
    \texttt{</first\_run\_reasoning\_traces>} \\
    \bottomrule
  \end{tabular}
\end{table*}

Here $P_{\mathcal D}$ is part of $T$, rather than part of the original task
prompt $x$; its exact value for each dataset is given verbatim below.
Separately returned visible answers remain outside $T$.  The five source
reasoning fields are taken in repeat-index order.  When applying the
50,000-character cap, any reasoning field longer than 50,000 characters is
replaced by its first 50,000 characters followed by \texttt{...}; block
formatting then strips leading and trailing whitespace.

\subsection{Model-Specific Details}

\paragraph{Qwen 3.7 Max reasoning instruction.}
For every evaluated Qwen 3.7 Max condition, we use the following fixed
\texttt{xhigh} instruction recommended for reasoning scenarios in the official
Qwen 3.7 Max evaluation guidance available when the experiments were run:

\begin{quote}
\small\ttfamily
Reasoning effort is set to xhigh. Please think carefully through the task,
validate key assumptions, consider plausible alternatives, and prioritize
correctness, consistency, and clarity in the final answer.
\end{quote}

Here \texttt{xhigh} names the recommended prompt text rather than a
provider-side \texttt{reasoning\_effort} value. For MRCRv2, the runner prepends
the instruction to the system message. The GraphWalks and NUB-1M runners
prepend it to their single user message. Within each dataset, the delivery
format is fixed across the evaluated conditions and does not change the
relative placement of $T$ and the long context.

\paragraph{Version of Deepseek V4 Pro.}
There are two models both called Deepseek V4 Pro released on 2026-4-24 and 2026-8-13 respectively. We use the 2026-4-24 version (also called Deepseek V4 Pro Preview) for our experiments.

\subsection{Dataset-Specific Details}

\paragraph{GraphWalks.}
For GraphWalks, $P_{\mathcal D}$ is the following exact text:

\begin{quote}
\small\ttfamily
Below are selected reasoning traces or trace tail windows from independent
first attempts on the same graph problem. They may contain mistakes. Use them
only as scratchpad hints, and verify against the graph.
\end{quote}

We also use a system prompt:

\begin{quote}
\small\ttfamily
You solve directed-graph algorithm problems. Use only the graph and operation
in the user message. Return exactly one visible line in this format: Final
Answer: [node1, node2]. Use [] for the empty set. Do not include any text before
or after that line.
\end{quote}

The runner also appends this exact answer-format instruction to the user
message:

\begin{quote}
\small\ttfamily
Return exactly one line in this format: Final Answer: [node1, node2]. Use [] for
the empty set.
\end{quote}

For $[T,x]$, the complete trace block precedes the complete released
GraphWalks prompt, and the answer-format instruction follows that prompt.  For
$[x,T]$, the runner splits the released prompt immediately before its last
\texttt{Operation:} block: the graph instructions and edge list come first,
then the identical trace block, then the final operation and answer-format
instruction.

This formatting instruction guides the model to output
its answer in a parse-able way. 

\paragraph{MRCRv2.}
For MRCRv2, $P_{\mathcal D}$ is the following exact text:

\begin{quote}
\small\ttfamily
Below are reasoning traces from independent first attempts on the same MRCR
problem. They may contain mistakes. Use them only as scratchpad hints; verify
against the conversation and final request. Do not copy any trace text into the
visible answer.
\end{quote}

\FloatBarrier
\section{GraphWalks Difficulty Profiles}
\label{app:graphwalk-difficulty-profile}

As an exploratory diagnostic, we examine whether the descriptive timing
gap varies with two observable GraphWalks properties. BFS specifies a requested
traversal depth $d$, and Parents has a gold parent-set size $k$. These variables
summarize aspects of traversal and aggregation demand, but realized difficulty
can also depend on frontier size, early termination, and graph structure. The
analysis and highlighted ranges were developed after inspecting outcomes; they
are hypothesis-generating rather than confirmatory tests of an interaction,
threshold, capacity limit, or mechanism.

We pool adjacent low-support values into BFS bins $1$--$2$, $3$--$4$,
$5$--$6$, $7$--$8$, and $9$--$10$, and Parents bins $0$, $1$, $2$, $3$,
$4$--$5$, and $\geq6$.  Their problem counts are $(17,25,21,15,22)$ and
$(13,17,25,20,13,12)$, respectively.  For feedback timing
$p\in\{\mathrm{pre},\mathrm{post}\}$ and bin $b$, we report the gain over
the common first-pass baseline,
\begin{equation}
  \Delta_p(b)=\operatorname{Score}_p(b)
  -\operatorname{Score}_{\mathrm{first}}(b).
  \label{eq:graphwalk-difficulty-delta}
\end{equation}
Each point averages five stored repeats within each problem and then the
problems in its bin. The yellow regions are descriptive and outcome-informed;
their boundaries are specific to each model, subtask, and metric. They were
not selected independently of the displayed scores and do not support
confirmatory inference.

\begin{figure*}[!t]
  \centering
  \includegraphics[width=0.85\textwidth]{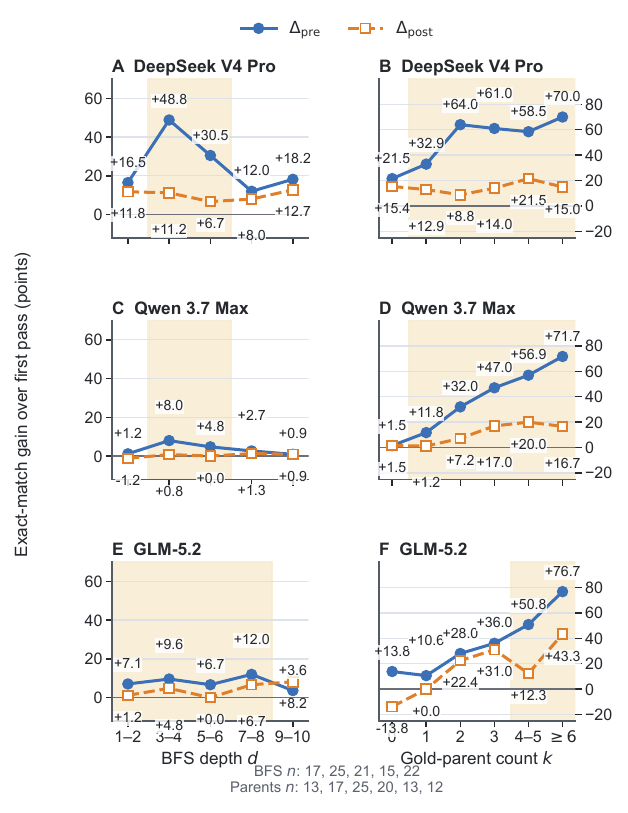}
  \caption{GraphWalks exact-match gains over the first pass by BFS depth $d$
  (left) and gold-parent count $k$ (right).  The blue and orange curves report
  $\Delta_{\mathrm{pre}}$ and $\Delta_{\mathrm{post}}$.  Bin supports appear
  below the panels.  Yellow shading marks descriptive, outcome-informed ranges
  of larger separation, with boundaries chosen separately for each model and
  subtask.}
  \label{fig:graphwalk-em-difficulty}
\end{figure*}

The prefix-minus-postfix exact-match gap in
Figure~\ref{fig:graphwalk-em-difficulty} is concentrated in particular bins.
DeepSeek V4 Pro has its largest BFS timing gaps
at $d=3$--$6$, and Qwen 3.7 Max has a smaller concentration in the same range.
GLM-5.2 has positive gaps through $d=8$ and a reversal in the final pooled
bin.  The Parents profiles differ.  \tas{} is never below \tappend{} in the plotted
exact-match bins, with a tie for Qwen 3.7 Max at $k=0$.  Qwen's separation
grows with $k$; DeepSeek V4 Pro and GLM-5.2 are nonmonotonic but have large
gaps in selected middle or high-$k$ bins.

\begin{figure*}[t]
  \centering
  \includegraphics[width=0.96\textwidth]{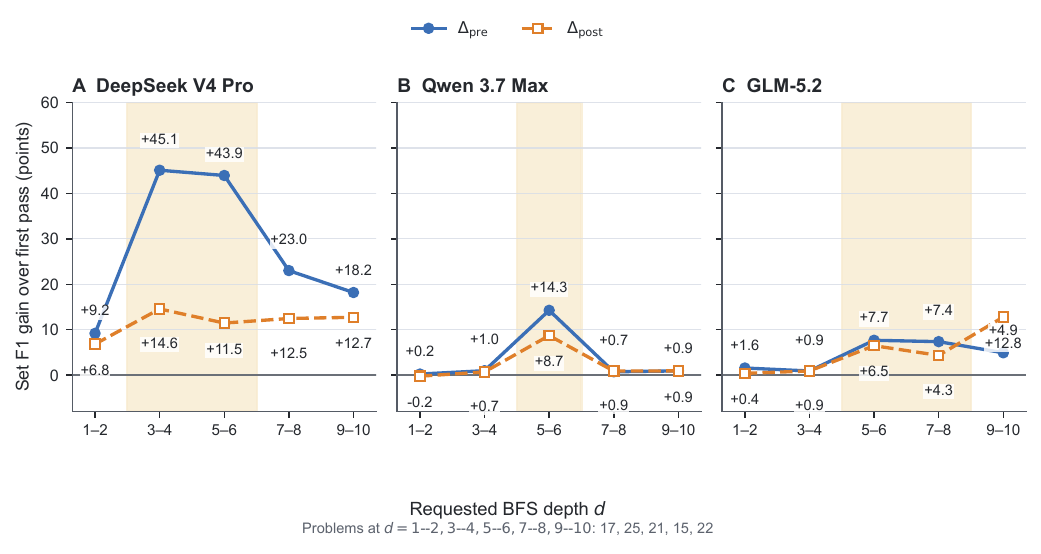}
  \par\vspace{0.35em}
  \includegraphics[width=0.96\textwidth]{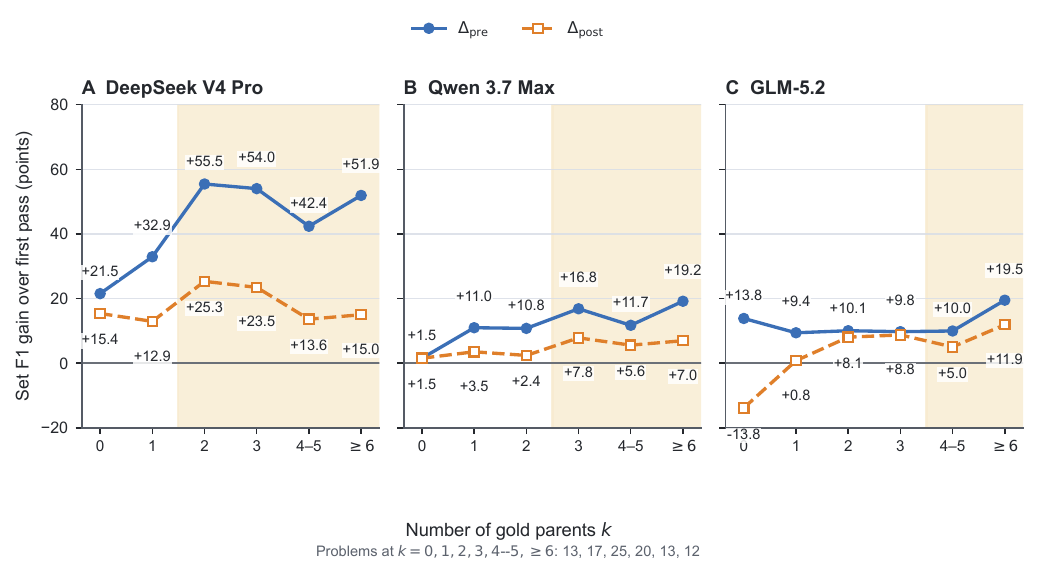}
  \caption{Set F1 gains for the same BFS (top) and Parents (bottom)
  stratifications. Set F1 records partial overlap between predicted and gold
  node sets.  Yellow shading again marks descriptive, outcome-informed ranges
  selected separately for each model, subtask, and metric.}
  \label{fig:graphwalk-f1-difficulty}
\end{figure*}

\clearpage

The binned set F1 means preserve the strong DeepSeek V4 Pro BFS pattern, while
Qwen 3.7 Max and GLM-5.2 show smaller or more localized separation. Parents
generally favors \tas{}, with substantial variation in the size of the gap.
These profiles document heterogeneity but do not establish why it occurs or
that depth or parent-set size mediates trace reuse. A confirmatory follow-up
would define bins or continuous contrasts before observing timing outcomes
and evaluate them on held-out problems.

\section{Confidence Intervals}
\label{app:bootstrap-confidence-intervals}

This appendix reports paired uncertainty for the principal same-\(T\)
timing contrast and cell-wise uncertainty for the broader results and
ablations. Table~\ref{tab:paired-placement-bootstrap} reports the mean
paired \tas{}-minus-\tappend{} difference and an
unadjusted 95\% percentile interval after averaging the five repeats within each
problem. The intervals lie fully above zero for 20 of 24 cells. The four
intervals that cross zero are DeepSeek V4 Pro MRCRv2 512K EM, Qwen 3.7 Max
GraphWalks BFS F1, and GLM-5.2 GraphWalks BFS EM and F1.

\begin{table}[t]
  \centering
  \scriptsize
  \setlength{\tabcolsep}{2.5pt}
  \renewcommand{\arraystretch}{0.90}
  \caption{Paired problem-cluster uncertainty for the strict same-\(T\) timing contrast. $\Delta$ is Trace as State minus Trace Append in percentage points; brackets give unadjusted 95\% percentile intervals over problems after five-repeat averaging ($n=100$ per cell; 20{,}000 resamples; seed 0). GW is GraphWalks 256K and MR is MRCRv2 8-needle. Intervals wholly above zero are bold.}
  \label{tab:paired-placement-bootstrap}
  \resizebox{\columnwidth}{!}{%
  \begin{tabular}{@{}llr@{}}
    \toprule
    Model & Cell & $\Delta$ [95\% CI] \\
    \midrule
    \multirow{8}{*}{\shortstack[l]{DeepSeek\\V4 Pro}} & GW BFS EM & $+17.00\;\mathbf{[+11.00,\,+23.40]}$ \\
     & GW BFS F1 & $+17.63\;\mathbf{[+11.65,\,+23.97]}$ \\
     & GW Par. EM & $+38.80\;\mathbf{[+31.60,\,+46.20]}$ \\
     & GW Par. F1 & $+26.02\;\mathbf{[+20.52,\,+31.84]}$ \\
    \cmidrule(lr){2-3}
     & MR 256K EM & $+10.20\;\mathbf{[+4.80,\,+15.80]}$ \\
     & MR 256K Seq. & $+9.11\;\mathbf{[+4.77,\,+13.93]}$ \\
     & MR 512K EM & $+3.20\;[-2.60,\,+9.40]$ \\
     & MR 512K Seq. & $+6.35\;\mathbf{[+1.69,\,+11.36]}$ \\
    \midrule
    \multirow{8}{*}{\shortstack[l]{Qwen 3.7\\Max}} & GW BFS EM & $+3.40\;\mathbf{[+1.00,\,+6.40]}$ \\
     & GW BFS F1 & $+1.29\;[-0.56,\,+3.90]$ \\
     & GW Par. EM & $+25.40\;\mathbf{[+17.60,\,+33.60]}$ \\
     & GW Par. F1 & $+7.43\;\mathbf{[+4.63,\,+10.58]}$ \\
    \cmidrule(lr){2-3}
     & MR 256K EM & $+4.40\;\mathbf{[+1.40,\,+8.00]}$ \\
     & MR 256K Seq. & $+4.19\;\mathbf{[+1.41,\,+7.53]}$ \\
     & MR 512K EM & $+6.40\;\mathbf{[+1.80,\,+11.40]}$ \\
     & MR 512K Seq. & $+5.11\;\mathbf{[+0.67,\,+9.84]}$ \\
    \midrule
    \multirow{8}{*}{GLM-5.2} & GW BFS EM & $+3.40\;[-1.00,\,+7.80]$ \\
     & GW BFS F1 & $-0.83\;[-3.94,\,+1.62]$ \\
     & GW Par. EM & $+16.80\;\mathbf{[+10.40,\,+23.80]}$ \\
     & GW Par. F1 & $+7.32\;\mathbf{[+3.93,\,+11.30]}$ \\
    \cmidrule(lr){2-3}
     & MR 256K EM & $+21.20\;\mathbf{[+14.40,\,+28.20]}$ \\
     & MR 256K Seq. & $+13.56\;\mathbf{[+8.33,\,+19.34]}$ \\
     & MR 512K EM & $+10.80\;\mathbf{[+3.40,\,+18.20]}$ \\
     & MR 512K Seq. & $+10.20\;\mathbf{[+4.65,\,+16.08]}$ \\
    \bottomrule
  \end{tabular}%
  }
\end{table}

\begin{figure*}[!p]
  \centering
  \includegraphics[width=0.72\textwidth]{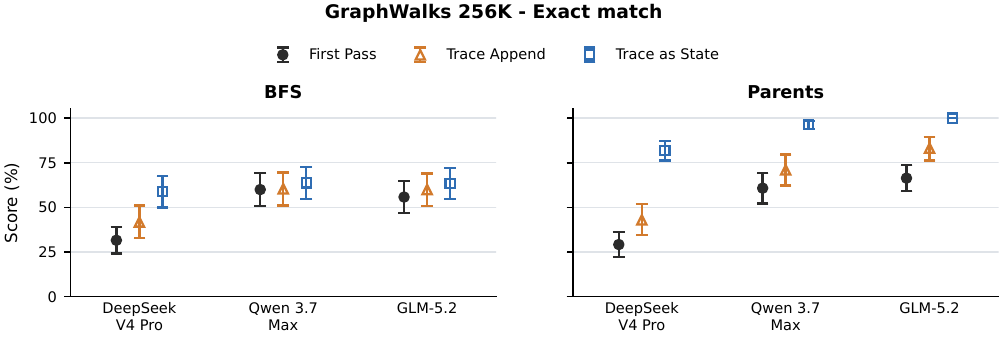}
  \par\vspace{0.05em}
  \includegraphics[width=0.72\textwidth]{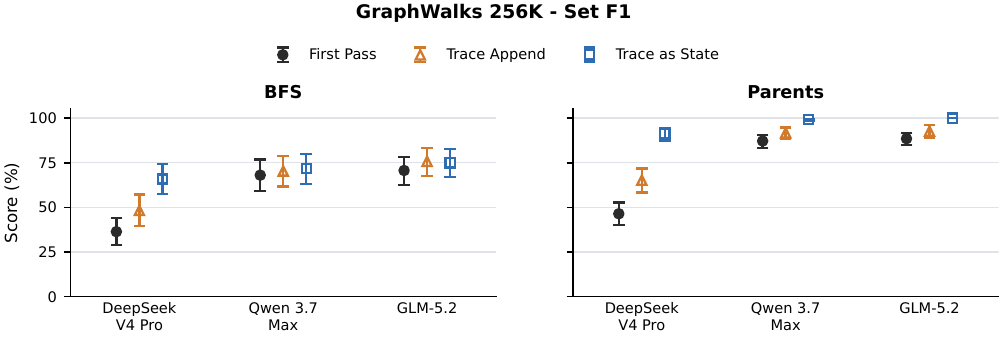}
  \par\vspace{0.05em}
  \includegraphics[width=0.72\textwidth]{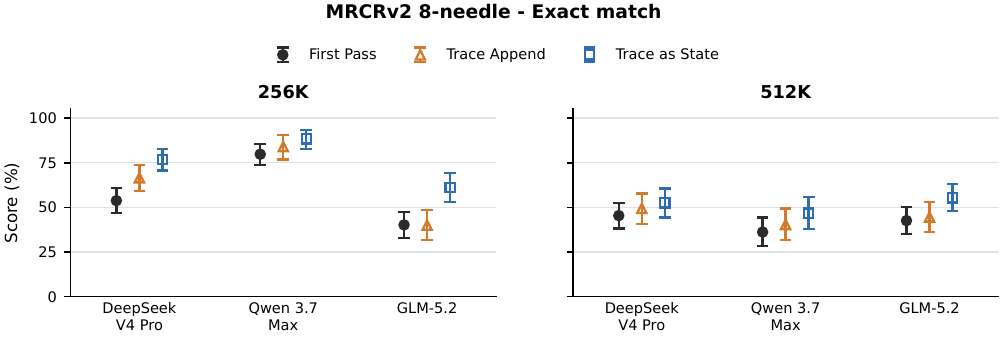}
  \par\vspace{0.05em}
  \includegraphics[width=0.72\textwidth]{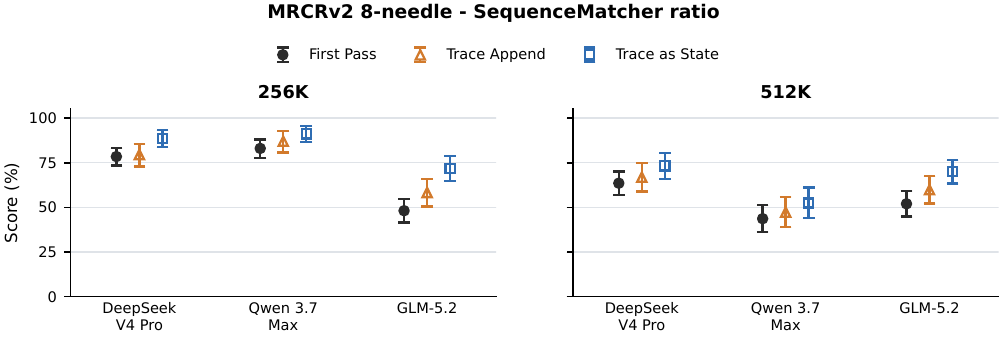}
  \par\vspace{0.05em}
  \includegraphics[width=0.32\textwidth]{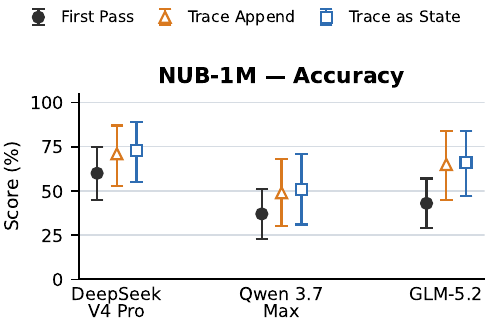}
  \caption{Uncertainty for the strict same-\(T\) comparisons. The top two rows show
  GraphWalks 256K exact match and set F1, split into BFS and Parents; the bottom
  two wide rows show MRCRv2 8-needle exact match and SequenceMatcher ratio,
  split into 256K and 512K bins. The final panel shows NUB-1M accuracy. Points
  are means; bars are 95\% percentile intervals after averaging 5 repeats
  within each problem.}
  \label{fig:bootstrap-main-strict}
\end{figure*}

\begin{figure*}[!t]
  \centering
  \includegraphics[width=0.86\textwidth]{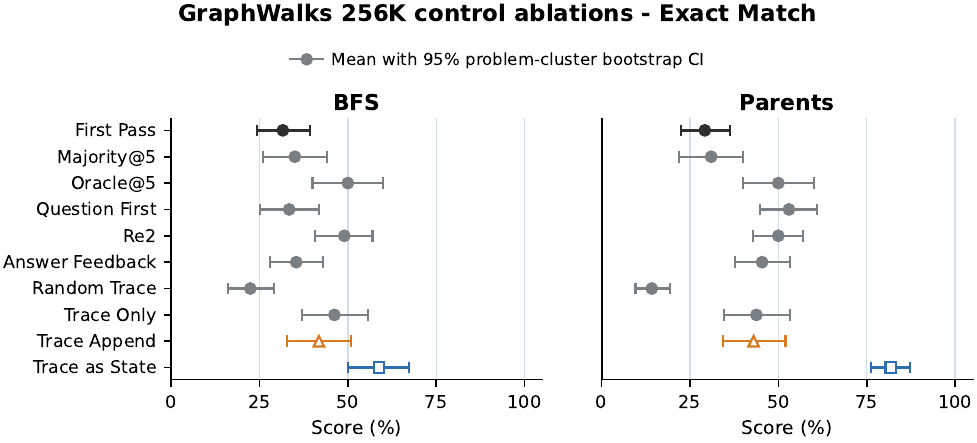}
  \par\vspace{0.15em}
  \includegraphics[width=0.86\textwidth]{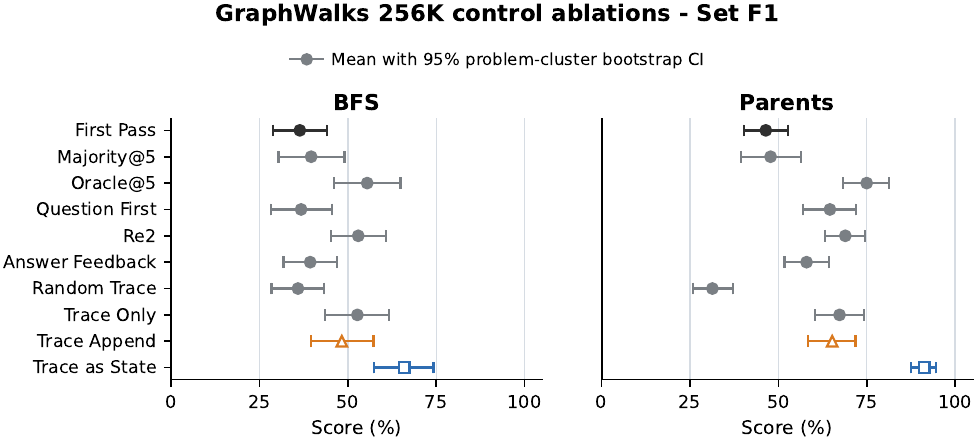}
  \caption{Uncertainty for DeepSeek V4 Pro GraphWalks 256K control comparisons
  in exact match and set F1. Points are means; bars are 95\% percentile
  intervals from 20{,}000 problem-cluster bootstrap resamples (seed~0).
  Majority@5 retains answer elements occurring in at least three repeat-level
  prediction sets, and Oracle@5 reports the retrospective maximum evaluator
  score among the five outputs. Other conditions average five planned
  repeats within each problem and assign zero to missing or invalid outputs.}
  \label{fig:bootstrap-supporting-controls}
\end{figure*}

Figures~\ref{fig:bootstrap-main-strict}
and~\ref{fig:bootstrap-supporting-controls} report cell means with 95\%
percentile intervals from the same problem-cluster bootstrap. Repeats are
averaged within each problem.

Figure~\ref{fig:bootstrap-main-strict} covers the strict same-\(T\) GraphWalks
and MRCRv2 comparisons in Table~\ref{tab:main-results}, together with the
supporting NUB-1M results. The NUB-1M intervals are wider, consistent with its
20 evaluated questions. Figure~\ref{fig:bootstrap-supporting-controls} covers
the GraphWalks control ablation. The same bootstrap procedure produces the
shaded trace-count intervals in Figure~\ref{fig:trace-count-ablation}.

\section{Token Usage}
\label{app:main-token-usage}

Table~\ref{tab:main-token-usage} reports token counts returned by provider
APIs for the solver calls underlying the main results.

For each retained response with provider-reported usage, we report cached
input, missed input, and output tokens. Missed input is total input minus cached input.
Reasoning tokens are included in the output.

 First Pass is the shared source pass used to construct \(T\).  For \tas{} and \tappend{}, we also report the
\emph{Total tokens} columns that add the corresponding First Pass counts component-wise.  

\begin{table*}[!t]
  \centering
  \scriptsize
  \setlength{\tabcolsep}{3.0pt}
  \renewcommand{\arraystretch}{0.93}
  \caption{Provider-reported token usage for main-result inference, in
  millions.  \emph{Tokens} reports the pass named in each row.
  For \tappend{} and \tas{}, \emph{Total tokens} adds First Pass
  component-wise; those columns are blank for First Pass itself.  Missed
  input is total input minus cached input.}
  \label{tab:main-token-usage}
  \begin{threeparttable}
  \begin{tabular}{@{}llrrrrrr@{}}
    \toprule
    Benchmark & Condition &
    \multicolumn{3}{c}{Tokens} &
    \multicolumn{3}{c}{Total tokens} \\
    \cmidrule(lr){3-5}\cmidrule(lr){6-8}
    & &
    \shortstack[r]{Cached\\input} &
    \shortstack[r]{Missed\\input} &
    Output &
    \shortstack[r]{Cached\\input} &
    \shortstack[r]{Missed\\input} &
    Output \\
    \midrule
    \multicolumn{8}{@{}l}{\emph{DeepSeek V4 Pro}} \\
    GraphWalks 256K & First Pass & 206.766 & 51.092 & 62.507 &  &  &  \\
     & \tappend{} & 271.017 & 68.554 & 35.231 & 477.783 & 119.646 & 97.738 \\
     & \tas{} & 277.074 & 62.501 & 27.975 & 483.840 & 113.593 & 90.482 \\
    \cmidrule(lr){1-8}
    MRCRv2 256K & First Pass & 50.817 & 46.961 & 1.149 &  &  &  \\
     & \tappend{} & 79.982 & 20.820 & 0.690 & 130.799 & 67.782 & 1.839 \\
     & \tas{} & 51.875 & 50.051 & 1.355 & 102.692 & 97.012 & 2.504 \\
    \cmidrule(lr){1-8}
    MRCRv2 512K & First Pass & 155.283 & 38.195 & 0.989 &  &  &  \\
     & \tappend{} & 178.675 & 18.526 & 0.702 & 333.958 & 56.721 & 1.691 \\
     & \tas{} & 157.261 & 39.940 & 1.241 & 312.544 & 78.135 & 2.230 \\
    \cmidrule(lr){1-8}
    NUB-1M Season 2 & First Pass & 31.651 & 8.418 & 0.817 &  &  &  \\
     & \tappend{} & 39.589 & 4.460 & 0.184 & 71.240 & 12.878 & 1.001 \\
     & \tas{} & 27.418 & 16.637 & 0.398 & 59.069 & 25.055 & 1.215 \\
    \midrule
    \multicolumn{8}{@{}l}{\emph{Qwen 3.7 Max}} \\
    GraphWalks 256K & First Pass & 260.213 & 97.483 & 5.402 &  &  &  \\
     & \tappend{} & 264.906 & 117.866 & 2.865 & 525.119 & 215.350 & 8.267 \\
     & \tas{} & 271.742 & 111.030 & 3.615 & 531.955 & 208.514 & 9.017 \\
    \cmidrule(lr){1-8}
    MRCRv2 256K & First Pass & 38.337 & 63.495 & 2.111 &  &  &  \\
     & \tappend{} & 60.227 & 51.195 & 0.989 & 98.564 & 114.690 & 3.100 \\
     & \tas{} & 62.213 & 49.209 & 1.397 & 100.550 & 112.704 & 3.508 \\
    \cmidrule(lr){1-8}
    MRCRv2 512K$^{\dagger}$ & First Pass & 134.607 & 63.738 & 2.263 &  &  &  \\
     & \tappend{} & 154.827 & 53.489 & 1.211 & 289.434 & 117.226 & 3.474 \\
     & \tas{} & 153.971 & 54.345 & 1.724 & 288.577 & 118.083 & 3.988 \\
    \cmidrule(lr){1-8}
    NUB-1M Season 2 & First Pass & 34.253 & 7.547 & 0.489 &  &  &  \\
     & \tappend{} & 38.184 & 5.773 & 0.253 & 72.437 & 13.320 & 0.742 \\
     & \tas{} & 32.414 & 11.549 & 0.419 & 66.667 & 19.096 & 0.908 \\
    \midrule
    \multicolumn{8}{@{}l}{\emph{GLM-5.2}} \\
    GraphWalks 256K & First Pass & 134.748 & 150.760 & 19.463 &  &  &  \\
     & \tappend{} & 55.612 & 265.396 & 9.127 & 190.361 & 416.156 & 28.591 \\
     & \tas{} & 64.390 & 256.618 & 16.491 & 199.138 & 407.377 & 35.954 \\
    \cmidrule(lr){1-8}
    MRCRv2 256K & First Pass & 73.782 & 24.718 & 1.891 &  &  &  \\
     & \tappend{} & 82.413 & 20.940 & 2.518 & 156.195 & 45.658 & 4.409 \\
     & \tas{} & 82.464 & 20.888 & 1.516 & 156.246 & 45.606 & 3.407 \\
    \cmidrule(lr){1-8}
    MRCRv2 512K & First Pass & 151.247 & 42.262 & 1.441 &  &  &  \\
     & \tappend{} & 157.740 & 40.508 & 2.002 & 308.987 & 82.770 & 3.443 \\
     & \tas{} & 157.458 & 40.790 & 1.648 & 308.705 & 83.052 & 3.089 \\
    \cmidrule(lr){1-8}
    NUB-1M Season 2 & First Pass & 1.272 & 41.123 & 0.995 &  &  &  \\
     & \tappend{} & 3.840 & 43.087 & 0.484 & 5.111 & 84.209 & 1.478 \\
     & \tas{} & 0.457 & 46.475 & 0.856 & 1.728 & 87.598 & 1.850 \\
    \bottomrule
  \end{tabular}
   \end{threeparttable}
\end{table*}

\section{Reproducibility}
We provide reference code and experimental records on \url{https://huggingface.co/datasets/Trace-as-state/Trace-as-State}. 

Some companies may obsolete their specific model endpoints, but open-source models are still downloadable to reproduce the results via self-hosted endpoints. 

The provided code also supports reproducing \tas{} on any potential future models that returns full reasoning traces, or on self-hosted open source models, as well as these models are based on causal transformers.

\end{document}